# MPIPN: A Multi Physics-Informed PointNet for solving parametric acoustic-structure systems

Chu Wang[1], Jinhong Wu[1], Yanzhi Wang[1], Zhijian Zha[1], Qi Zhou[1,*]

1. School of Aerospace Engineering, Huazhong University of Science & Technology, Wuhan, Hubei, 430074, China

**Abstract**

Machine learning is employed for solving physical systems governed by general nonlinear partial differential equations (PDEs). However, complex multi-physics systems such as acoustic-structure coupling are often described by a series of PDEs that incorporate variable physical quantities, which are referred to as parametric systems. There are lack of strategies for solving parametric systems governed by PDEs that involve explicit and implicit quantities. In this paper, a deep learning-based Multi Physics-Informed PointNet (MPIPN) is proposed for solving parametric acoustic-structure systems. First, the MPIPN induces an enhanced point-cloud architecture that encompasses explicit physical quantities and geometric features of computational domains. Then, the MPIPN extracts local and global features of the reconstructed point-cloud as parts of solving criteria of parametric systems, respectively. Besides, implicit physical quantities are embedded by encoding techniques as another part of solving criteria. Finally, all solving criteria that characterize parametric systems are amalgamated to form distinctive sequences as the input of the MPIPN, whose outputs are solutions of systems. The proposed framework is trained by adaptive physics-informed loss functions for corresponding computational domains. The framework is generalized to deal with new parametric conditions of systems. The effectiveness of the MPIPN is validated by applying it to solve steady parametric acoustic-structure coupling systems governed by the Helmholtz equations. An ablation experiment has been implemented to demonstrate the efficacy of physics-informed impact with a minority of supervised data. The proposed method yields reasonable precision across all computational domains under constant parametric conditions and changeable combinations of parametric conditions for acoustic-structure systems.

* Corresponding author: qizhou@hust.edu.cn & qizhouhust@gmail.com (Qi Zhou)

# 1 Introduction

Acoustic-structure systems are classic and vital multi-physics systems that can be abstracted from various engineering scenarios such as acoustic focusing [1], sonar cloaking [2], and sound absorption [3]. Mastering the physical performance of acoustic-structure systems is fundamental to acoustic structure design. Modeling and forecasting the dynamics of multi-physics systems remains an open scientific problem [4]. Solving parametric multi-physics systems from the mechanism necessitates designing a general framework that can identify and solve components of systems governed by corresponding governing equations. Commonly, the governing equations for the dynamics of physical systems are PDEs [5]. Deep learning methods, especially the neural network architecture have become a hotspot because of their ability in high-dimensional nonlinear mapping and automatic extraction of features. In the field of PDE solving, deep learning methods of physics-informed models were first used by Raissi et al. [6] to solve both linear and nonlinear PDEs in forward and inverse forms. It now has been broadly used to deal with various physical issues such as heat transfer [7], mechanics [8], and fluid dynamics [9]. For physics-informed methods, physical information is embedded in three pathways [10] separately or in tandem to accelerate training and enhance generalization. First, the utilization of variable fidelities observations that provide monitor data across the physical spatial and temporal field as supervised regime [11]. Wu et al. [12] constructed a multi-fidelity neural network to predict the acoustic pressure through data of multiple precision. Second, specialized architectures of neural networks are designed to embed the priori physical knowledge into the architecture itself. PointNet [13] extracts both local and global depth of field information to realize segmentation tasks. Kashefi et al. [14] designed a framework based on PointNet [13] to solve fluid flow fields on irregular geometries. Lagaris [15] proposed a systematic construction of neural networks to satisfy the initial boundary conditions and interface conditions of differential equations. Gao et al. [16] constructed a deep auto-encoder network to predict the peak points and directed the design of the Helmholtz resonator. Wang et al. [17] and Liu et al. [18] encoded multiple scale features to solve oscillatory stokes flows and Poisson–Boltzmann equations. Third, imposing physical constraints into the training process such as penalizing the residual of governing equations in loss function [6]. These aim to incorporate any domain-based physical knowledge into machine learning models in a flexible manner [19]. More related work can be seen in Refs.[20-22].

Furthermore, dedicating to constant computational conditions is insufficient to fully characterize performances of parametric physical systems [23]. Taking the parametric acoustic-structure systems as examples, changeable physical quantities including frequency, modulus and density affect solutions of governing PDEs. To broaden the applications of deep learning methods modeling parametric physical systems, there is a need for advanced solvers tailored for parametric PDEs that govern those physical systems. Im et al. [24, 25] used evolutionary processes to randomly generate diverse PDEs based on priori data of a multi-physics system. Berner et al. [26] solved numerical solutions of the parametric families of high-dimensional linear Kolmogorov PDEs by transforming multiple PDEs of interest to a single statistical learning problem. Chen et al. [27] used meta-learning to solve the parametric PDEs as a multi-task learning problem and guaranteed convergence for Poisson's equation. Isogeometric neural networks [28] are proposed to combine the solutions of PDEs with physical domains as linear Isogeometric representation and approximate based on domain features. PhyGeoNet [29] used CNN architecture by elliptic coordinate mapping from the irregular physical domain and regular reference domain. Kashefi et al. [30] proposed PIPN to solve the PDEs on multiple computational domains with irregular geometries. Operator learning methods were first introduced by DeepONet [31-33], which consists of two networks. The branch network deals with input function, while the truck network connects the output function at selected observation points. Li et al. [34] mapped families of Burgers' equation, Darcy flow, and Navier-Stokes equation into the frequency domain space to impose constraints by parameterizing the Fourier operator to embed in neural networks. Wavelet Neural Operator [35] further generalizes the neural operator by inducing spectral decomposition-based operator. In summary, physics-informed PDE solvers aim to come up with fast prediction mechanisms that are embedded with the physical priori knowledge to solve PDEs of interest.

However, mechanisms in the aforementioned cited work are characterized by two unresolved issues. First, data-driven methods derive the solutions of systems by being trained with massive observation data that are extensively designed manually [23-26, 31-33]. This requires generating labeled data coming at a strenuous consumption of time and ignores the interrelated mathematical and spatial information. Once the model is trained under a constant parametric condition, it is unable to fit in with new parametric conditions of systems. Second, in certain methods [27-30], parametric conditions are encoded as the input sequence to differentiate the parametric systems. On the one hand, the discretization of systems is elusive to be captured. On the other hand, the interrelated correlations between parametric conditions and computational domains fail to be represented. Therefore, models are incapable of simultaneously identifying parametric conditions and spatial coordinates in parametric systems that contain multiple computational domains. In the parametric acoustic-structure systems, combinations of diverse physical quantities and multiple computational domains make these two problems inevitable.

In order to address the aforementioned issues, we propose a Multi Physics-informed PointNet architecture, referred to as the MPIPN, to solve parametric acoustic-structure systems. The parametric systems include explicit physical quantities that are directly formulated in the governing PDEs and implicit quantities that indirectly affect the solutions. The combinations of these quantities form the parametric conditions of systems. Based on point-cloud architecture, the MPIPN deals with overall parametric conditions, and calculates pointwise solutions of governing PDEs for systems. The MPIPN is trained for each computational domain in systems through the compositions of corresponding governing PDEs and a minority of observation solutions as weakly supervised. The proposed method is utilized to solve the steady parametric acoustic-structure systems [36] governed by the Helmholtz equations. Two discrete acoustic-structure systems generate three numerical cases to test the efficiency of the MPIPN. The salient contributions of this paper are as follows:

(1) The MPIPN utilizes enhanced point-cloud architecture to directly compute the Helmholtz equations at discrete points and learns the correlations of spatial coordinates. It circumvents the necessity of massive manually designing supervised learning. Indeed, less than 3% of total discrete points are required for observation solutions while training to reasonable accuracy.

(2) The MPIPN preserves the constitutive mathematical information of acoustic-structure systems through constructing physics-informed loss function to calculate the values of parametric PDEs directly by the neural networks. Every computational domain in the system can be separately identified and solved. This ensures the scalability of the model to new parametric conditions without inducing extra labeled data.

(3) The MPIPN embeds and fuses explicit and implicit parametric conditions with spatial coordinates of multiple computational domains. The correlation of parametric conditions and spatial features can be simultaneously extracted to identify a series of parametric PDEs governing different domains in acoustic-structure systems. Multiple computational conditions can be identified and integrated to solve.

The rest of this paper is organized as follows: The background of PDEs for parametric physical systems and point-cloud based architecture for PDE solvers are discussed in Sect. 2. In Sect. 3, we introduce the methodology of the proposed framework MPIPN, including its architecture and adaptive physics-informed loss function. In Sect. 4, numerical experiments of acoustic-structure systems are implemented by applying the MPIPN to test its effectiveness in solving the governing PDEs. Three cases including constant explicit and implicit conditions and changeable combinations of physical conditions are executed as validation as well as the ablation experiment. Finally. Sect. 5 discusses the conclusions and future works of MPIPN.

## 2 Background

### 2.1 PDEs of parametric systems

Real-world physical effects can be abstracted into different multi-physics systems. Solving physical systems is solving corresponding governing equations at essence. In physical systems, PDEs can describe a wide range of physical phenomena and dynamics, including heat transfer, convection-diffusion processes, fluid dynamics, etc. A multi-physics system is usually composed of several types of physical fields and coupling relationships, including fluid-structure coupling, thermo-structure coupling, acoustic-structure coupling, etc. In such multi-physics systems, there are multiple computational domains with independent governing PDEs. This can be mathematically represented as

$$S\left[(u(t,\mathbf{x}))\right] = \{D_1[u(t,\mathbf{x})],...,D_i[u(t,\mathbf{x})],...,D_n[u(t,\mathbf{x})]\}, \tag{1}$$

where $S(\cdot)$ denotes the governing PDEs for the system, $u(t,\mathbf{x})$ is the solution, and $D_i(\cdot)$ is the governing equations for the $i$ th computational domain.

For a certain computational domain in physical systems, changeable systematic parameters cannot be characterized by a single fixed PDE. Parametric PDE is a series of PDEs that govern a certain computational domain with variable parameters. A general form of time-dependent governing parametric PDE can be written as

$$F\left(u, \frac{\partial u}{\partial t}, \frac{\partial u}{\partial \mathbf{x}}, \frac{\partial^2 u}{\partial t^2}, \frac{\partial u}{\partial \mathbf{x}^2}, \cdots\right) = f(t,\mathbf{x}), t \in [0,T], \mathbf{x} \in \Omega, \tag{2}$$

where $f(t,\mathbf{x})$ is the source term of PDE, $\mathbf{x} = (x_1, x_2, \cdots, x_n)$ is spatial vector, and $\Omega$ is the computational domain of spatial vectors. $F(\cdot)$ means nonlinear parametric mapping of governing equations. Since the system is time-dependent, a general form of initial condition can be written as

$$G\left(u, \frac{\partial u}{\partial t}, \frac{\partial^2 u}{\partial t^2}, \cdots\right) = g(\mathbf{x}), t = 0, \mathbf{x} \in \Omega, \tag{3}$$

where $g(\mathbf{x})$ is the source term of the initial condition PDE and $G(\cdot)$ is the nonlinear parametric mapping of the initial condition. Meanwhile, the boundary condition for time-dependent PDE can be written as

$$H\left(u, \frac{\partial u}{\partial t}, \frac{\partial u}{\partial \mathbf{x}}, \frac{\partial^2 u}{\partial t^2}, \frac{\partial^2 u}{\partial \mathbf{x}^2}, \cdots\right) = h(t, \mathbf{x}), t \in [0, T], \mathbf{x} \in \partial\Omega, \tag{4}$$

where $h(t, \mathbf{x})$ is the source term of boundary condition PDE, $\partial\Omega$ denotes the boundary of computational domain, and $H(\cdot)$ is the nonlinear parametric mapping of boundary condition. It should be noted that the nonlinear parametric mappings $F(\cdot)$, $G(\cdot)$ and $H(\cdot)$ contain physical quantities that remain variables in the computation process, from which all differential terms are subjected to the parametric PDEs family. For a computational domain with variable physical quantities, the parametric PDE family can characterize it as a collection. Some of the quantities exist as variables in the governing PDEs for the system, while others exist in the associated equations of the system other than directly in the governing equations. For example, the mapping function of three-dimensional non-source heat equation can be written as

$$F[u(t, \mathbf{x})] = u_t - \alpha\left(u_{xx} + u_{yy} + u_{zz}\right), \tag{5}$$

where $\alpha$ is the thermal diffusivity that exists as a variable in the governing PDE. Therefore, the mapping function $F[u(t, \mathbf{x})]$ represents the PDE family of three-dimensional non-source heat equation. In this paper, we solve the two-dimensional steady parametric acoustic-structure systems governed by parametric Helmholtz equations that contain both explicit and implicit physical quantities.

## 2.2 Point-cloud architecture

In this paper, point-cloud data is used to represent the spatial state of acoustic-structure systems. Point-cloud data is widely used in segmentation tasks. The common point-cloud is constructed with spatial coordinates of the target multidimensional domain. Taking the PointNet [13] for example, it utilizes the spatial information of point-cloud, and extracts local and global features to segment the images. The architecture of PointNet is shown in **Fig. 1**. Specifically, the number of points that constitute point-cloud is $N$, $d_{\text{input}}$, $d_{\text{output}}$ represents the input and output dimensions, respectively. Dimensions of hidden layers in PointNet are denoted as $N_{local}$, $N_1$, $N_1$ and $N_{global}$. When being applied to solve PDEs, by mapping the input of point-cloud to the output as the solutions under the constraints of PDE, PointNet can capture geometric features of the computational domains. The input are coordinates of point-cloud and the output are the solutions of each point in PDE. Therefore, PointNet can process changeable geometries of computational domains. Under the circumstance that different computational domains obey the same PDE with constant parametric conditions, the networks are first trained on a set of irregular geometries of computational domains $\Phi = \{V_i\}_{i=1}^{m}$. Then, the well-trained model is used for an unseen set of irregular geometries of computational domains $\Psi = \{V_i\}_{i=1}^{l}$.

There are three advantages that point-cloud architecture can provide. First, based on spatial discretization, point-cloud data can be extracted from vertices of structured or unstructured grids of the computational domains. The input of networks are sets of discretized spatial coordinates that explicitly represent the initial computational domains. This realizes computing the derivative terms on each point in PDEs directly with respect to the predicted solutions. Hence, we can judge and measure how well the current solution satisfies the PDEs. This is fundamental to the unsupervised or weakly supervised learning of physics-informed machine learning, since the priori solutions are not completely known. Second, the spatial information of coordinates can represent irregular geometries and describe computational domains. Unlike CNN which utilizes pixel data, the point-cloud data is suitable to be transformed in spatial dimension. The formation of the computational domain is determined by the spatial properties of the point-cloud, avoiding binary one-hot encoding. Third, the governing PDEs of the computational domain are trainable under the point-cloud architecture. a fixed PDE can be explicitly computed with respect to the point-cloud as a loss function for the training of the neural networks.

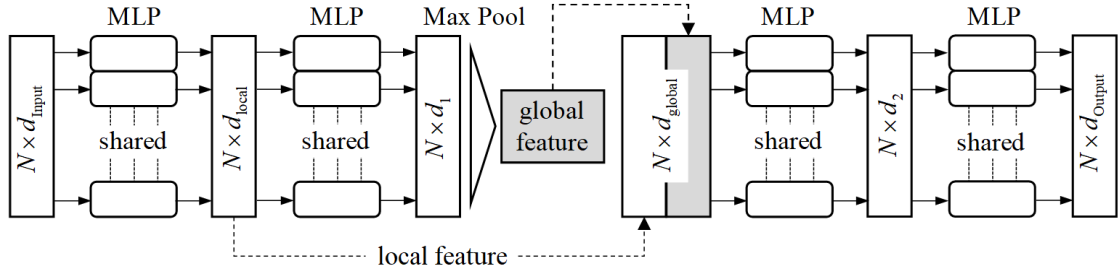

**Fig. 1** PointNet architecture for PDE solving

# 3 Methodology
## 3.1 Architecture of MPIPN

Our goal is to train the framework over the training dataset $\Gamma$ and use the well-trained model to solve new parametric conditions in the testing dataset $\Lambda$. The $\Gamma$ consists of the prior knowledge of the systems, while the $\Lambda$ contains points to be solved and parametric conditions. Specifically, the prior knowledge consists of four types of physical information to identify different computational domains and parametric conditions of the systems. First, the spatial point-cloud is extracted from discrete physical systems. Point-cloud represents geometric information and is fundamental to gradient-based computation. The mapping between the discrete point-cloud and the physical field numerically forms the solutions of the systems. Second, there are physical quantities that remain variables within the governing PDEs of the system, i.e., frequency, temperature, density and time series. Changeable combinations of physical quantities constitute parametric conditions for systems. These quantities directly or indirectly influence the solutions of PDEs. Therefore, physical quantities that appear directly in the PDEs are referred to as explicit quantities or explicit conditions. Physical quantities that cannot be explicitly expressed in the PDEs, but affect the solutions of the systems, are termed implicit quantities or implicit conditions. Third, the governing PDEs for corresponding computational domains of the systems. For a multi-physics system, each computational domain satisfies a particular PDE. Therefore, the constraints of the governing PDEs are supervised guidelines as the physics-informed information for training the framework. Fourth, the priori exact solutions as the observation values that rectify the predicted solutions to avoid the degenerate solutions. This is either detected by engineering sensors or provided by mathematically known solutions. Observation solutions should be randomly or naturally selected without deliberate design. For the testing dataset $\Lambda$, the former three types of physical information are included, while the priori solutions are prohibited when being used to solve unseen parametric conditions. The $\Lambda$ provides spatial, explicit and implicit physical quantities as input of the MPIPN and overall pointwise solutions of the systems are obtained as output. The overarching framework of the proposed is illustrated in **Fig. 2**, and details of substructures are shown in **Fig. 3**.

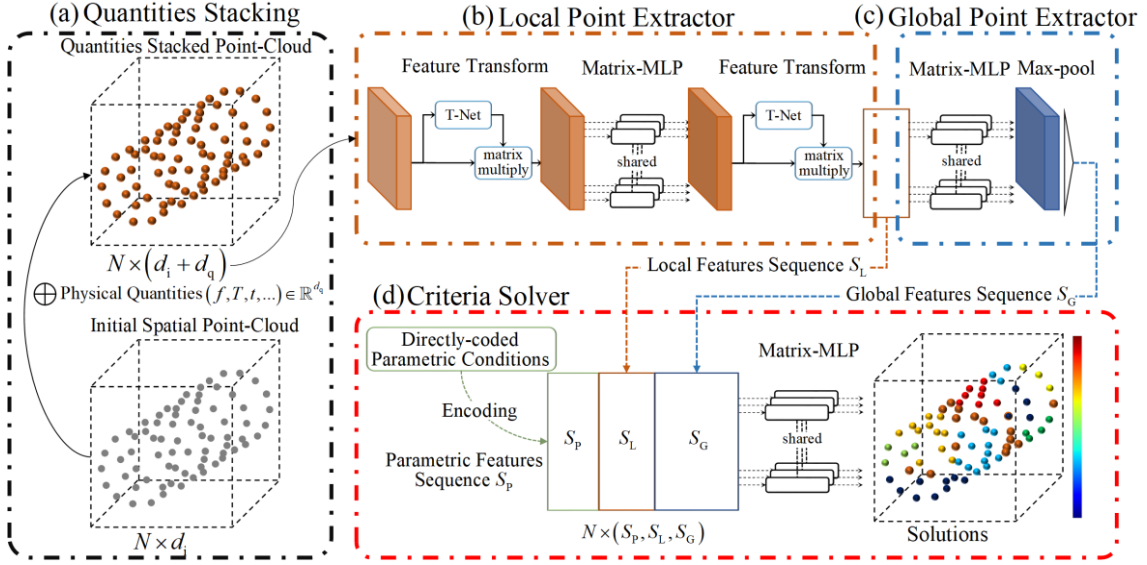

**Fig. 2** Multi Physics-Informed PointNet architecture with four modules

In general, the MPIPN receives spatial coordinates of the point-cloud and stacks with explicit physical quantities to obtain reconstructed point-cloud through the Quantities Stacking module. Then, Local Point Extractor and Global Point Extractor networks step-wisely extract the interrelated features of the spatial coordinates and explicit physical quantities derived from stacked point-cloud. Thus, we obtain the local features sequence and global features sequence, respectively. Afterward, the implicit physical quantities are encoded as parametric features to form solving criteria with local and global features sequences for the parametric PDEs on each discrete point. Finally, the Criteria Solver networks take solving criteria sequences as input and output pointwise solutions. The training of the architecture requires one-time gradient backpropagation for each iteration to update overall parameters of the neural networks.

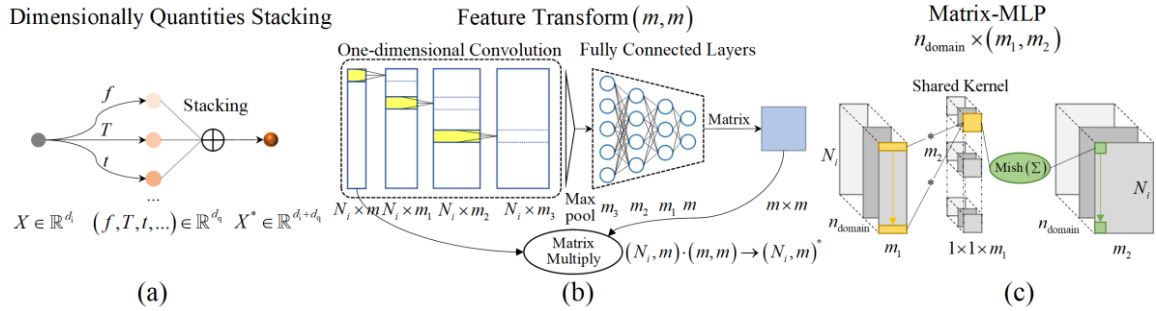

**Fig. 3** Details of substructures in the MPIPN: (a) Dimensionally Quantities Stacking; (b) Feature Transform method; (c) Matrix-MLP structure

The Quantities Stacking module is the first step of the framework. Based on the finite and boundary element (FE-BE) methods, the systems can be arbitrarily discretized. Therefore, spatial point-cloud can be extracted from the vertices of the meshes. We denote point-cloud as $X = \{x_i\}_{i=1}^{N} \in \mathbb{R}^{d_i}$ and explicit quantities as $Q = \{q_i\}_{i=1}^{N} \in \mathbb{R}^{d_q}$ in the $\Gamma$, where $N = \{N_i\}_{i=1}^{n_{domain}}$ indicates the total number of points for each computational domain, $n_{domain}$ denotes the number of computational domains within the systems, $d_i$ indicates the dimension of the initial spatial point-cloud. **Fig. 3 (a)** shows the Quantities Stacking module, through which, the initial spatial point-cloud and explicit quantities can be fused mathematically as

$$x_i^* = \left\{ x_i \in \mathbb{R}^{d_i} \oplus q_i \in \mathbb{R}^{d_q} \right\} \in \mathbb{R}^{d_i + d_q}, \tag{6}$$

where the reconstructed point-cloud $X^* = \{x_i^*\}_{i=1}^{N} \in \mathbb{R}^{d_i + d_q}$ is stacked dimensionally and contains pointwise spatial and explicit physical information.

The Local Point Extractor and Global Point Extractor both extract the features of the stacked point-cloud to capture the interrelation of the spatial and explicit quantities. Due to different sizes of receptive fields, global features represent the overall geometric characteristics and identify computational domains, while local features represent spatial correlation and depict gradient information in small range. The two extraction networks generate features of local and global scales by using Feature Transform and Matrix-MLP structures. **Fig. 3 (b), (c)** show the Feature Transform method and Matrix-MLP structure, respectively. In the Feature Transform method, the T-Net is realized by the composition of one-dimensional convolution and fully connected layers. Suppose that there are $N_i$ points in a reconstructed point-cloud with dimension of $m$ of the $i$ th computational domain. The input $(N_i \times m)$ is mapped by the T-Net to the transform matrix $(m \times m)$. Through the Matrix Multiply, we obtain $(N_i \times m)^*$ as the result of matrix dot product of the input and transform matrix. This procedure can be expressed as

$$(N_i \times m)^* = \mathcal{F}(N_i \times m) = (N_i \times m) \odot T(N_i \times m)_{m \times m}, \tag{7}$$

where, $\mathcal{F}(\cdot)$ denotes the overall mapping of Feature Transform $(m \times m)$, $T(\cdot)$ denotes the mapping of T-Net to a matrix in the shape of $m \times m$, $\odot$ represents matrix dot product. In Matrix-MLP structure noted as $n_{domain} \times (m_1, m_2)$, the input $(N_i \times m_1)$ of the $i$ th computational domain with dimension of $m_1$ is scanned by number of $m_2$ shared kernel of shape $1 \times 1 \times m_1$, and generates output $(N_i \times m_2)$. This can be mathematically expressed as

$$\{\mathbf{O}\}_{j=1}^{m_2} = \mathcal{M}\left(\{\mathbf{I}\}_{l=1}^{m_1}\right) = \text{Mish}\left(\sum_{l=1}^{m_1}\{k_l\}_{j=1}^{m_2} \cdot \mathbf{I}_l\right), \tag{8}$$

where $\mathcal{M}(\cdot)$ denotes the overall mapping of Matrix-MLP, $\mathbf{O}$ denotes the $j$ th dimension sequence of output with the shape of $N_i \times 1$, $k_l$ denotes the $l$ th element of the $j$ th kernel, $\mathbf{I}_l$ denotes the $l$ th dimension sequence of input with the shape of $N_i \times 1$, $\text{Mish}(\cdot)$ represents the Mish activation function [37]. The non-monotonic property of Mish causes small negative inputs to be preserved as negative and the order of continuity being infinite. The Mish activation function can be written as

$$\sigma(x) = x \cdot \tanh\left(\ln\left(1+e^x\right)\right), \tag{9}$$

where $\tanh(\cdot)$ is an activation function that can be written as

$$\tanh(z) = \frac{e^z - e^{-z}}{e^z + e^{-z}}. \tag{10}$$

The Feature Transform method achieves the transformation alignment, ensuring the point-cloud is robust in geometrical configuration. While, the Matrix-MLP structure extracts the features of the point-cloud automatically. Through the above two methods, local features sequence $S_L$ can be obtained as

$$S_L = \mathcal{M}\left(\mathcal{F}\left(x_i^*\right)\right). \tag{11}$$

The sequence is obtained by several connected blocks consisting of the Feature Transform and Matrix-MLP. The Global Point Extractor decodes the global features from the local features by utilizing permutation invariant function, max-pool. The global features sequence $S_G$ can be mathematically derived as

$$S_G = \max_j \left(x_{N_i}^j\right), \text{ with } i = 1, \cdots, n_{\text{domain}}; \ j = 1, \cdots, m_3, \tag{12}$$

where $m_3$ denotes the number of global features, and $x_{N_i}^j$ represents the $j$ th feature of input sequence on the $i$ th computational domain. Therefore, physical features are obtained as local features sequence $S_L$ and global features sequence $S_G$ by features extraction of various scales with the constructed point-cloud as input.

For those implicit parametric conditions, they are separately encoded by auto-encoder or statistically as parametric features sequence $S_P$ as

$$S_P = \mathcal{E}\left(q_i^*, g\left(q_i^*\right)\right), \text{ with } q_i^* \in \mathbb{R}^{d_q^*}, \tag{13}$$

where $Q^* = \{q_i^*\}_{i=1}^{N} \in \mathbb{R}^{d_q^*}$ denotes the implicit physical quantities, $g(\cdot)$ denotes the priori correlation of different implicit quantities, and $\mathcal{E}(\cdot)$ denotes the selected encoder mapping method. Finally, the solving criteria of parametric systems can be expressed as the concatenating of the above three sequences: $N \times (S_\text{P}, S_\text{L}, S_\text{G})$. Predicted solutions $Y = \{y_j\}_{j=1}^{N} \in \mathbb{R}^{d_s}$ are mapped nonlinearly by the solving criteria as

$$Y_i = \mathcal{C}_i\left(N_i \times (S_\text{P}, S_\text{L}, S_\text{G})_i\right), \quad i = 1, 2, \ldots, n_{\text{domain}} \tag{14}$$

where $\mathcal{C}_i(\cdot)$ denotes the $i$ th mapping from the solving criteria for individual computational domains to the solutions in the solver. Based on the corresponding solving criteria, parametric physical systems with multiple quantities can be solved.

## 3.2 Adaptive physics-informed loss function

In MPIPN, the mapping neural networks and encoder are trained simultaneously by PyTorch automatic gradient computation mechanism. All mapping functions including $\mathcal{F}(\cdot)$, $T(\cdot)$, $\mathcal{M}(\cdot)$, and $\mathcal{E}(\cdot)$ are trained by compositing the residuals of the governing PDEs and residuals of the predicted solutions with priori exact solutions at the observation points. Since the physical systems are discretized by the FE-BE method and the vertices of meshes are extracted, the values of the PDEs can be directly calculated. To solve the PDEs corresponding to multiple computational domains, the calculated values of the equations by predicted solutions on each domain should satisfy the equilibrium of the PDEs. This offers the inspiration of physics-informed loss function $\mathcal{L}_{\text{PDE}}$ for arbitrary computational domains as

$$\mathcal{L}_{\text{PDE}} = \frac{1}{N_\text{P}} \sum_{i=1}^{N_\text{P}} \left\| F\left(x_1, x_2, \ldots, x_n, \tilde{u}, \frac{\partial \tilde{u}}{\partial x_1}, \ldots \frac{\partial \tilde{u}}{\partial x_n}, \frac{\partial^2 \tilde{u}}{\partial x_1 \partial x_2}, \ldots, \frac{\partial^\alpha \tilde{u}}{\partial x^{\alpha_1}_1 \partial x^{\alpha_2}_2 \ldots \partial x^{\alpha_n}_n}\right)_i \right\|_\text{L}, \tag{15}$$

where $F(\cdot)$ is the function that represents the general PDE, $N_\text{P}$ denotes the number of points in the computational domain, $x_n$ denotes the variable quantities and $\tilde{u}$ denotes the predicted solutions on the $i$ th computational domain, $\|\cdot\|_\text{L}$ denotes the $L$ norms. By constructing the mean squared errors of the priori observation solutions, the observation loss function $\mathcal{L}_{\text{OBS}}$ can be written as

$$\mathcal{L}_{\text{OBS}} = \frac{1}{N_\text{o}} \sum_{j=1}^{N_i} \|\tilde{u} - u\|_\mathbf{2}, \tag{16}$$

where $N_o$ denotes the number of observation solutions, $u$ denotes the ground true solutions and $\|\cdot\|_2$ is the $L_2$ norm. It should be noted that $\mathcal{L}_{\text{PDE}}$ and $\mathcal{L}_{\text{OBS}}$ can be adaptively combined to generate suitable loss functions for multi-physics systems. Indeed, residuals of physics-informed loss function $\mathcal{L}_{\text{PDE}}$ and observation loss function $\mathcal{L}_{\text{OBS}}$ may be out of scale during the training process. To avoid the inconsistency of the scales between the two components of the loss function, an adaptive physics-informed loss function is proposed here. We first assigned a weight to each loss function component to balance all components to the same scale, and second to sum up balanced components to obtain the adaptive weighted loss function. The form of adaptive physics-informed loss function for an arbitrary domain can be mathematically written as

$$\mathcal{L} = \alpha \cdot \mathcal{L}_{\text{OBS}} + \beta \cdot \mathcal{L}_{\text{PDE}}, \tag{17}$$

where $\mathcal{L}_{\text{PDE}}$ indicates arbitrary loss function of discrete PDEs, $\mathcal{L}_{\text{OBS}}$ indicates the observation loss function domain, $\alpha$ and $\beta$ are corresponding learnable coefficients for each component. Here we fix $\alpha = 1$ and $\beta = 0.1$ to simplify the training.

## 4 Numerical experiments for acoustic-structure problems

In this section, the MPIPN is applied to solve two-dimensional acoustic-structure systems by obtaining pointwise scattered acoustic pressure. The system contains three computational domains with different governing PDEs. The numerical experiments are implemented to validate superiority of the MPIPN in solving constant and changeable parametric conditions of the Helmholtz equations of the parametric systems.

## 4.1 Application of MPIPN to acoustic-structure systems

### 4.1.1 Acoustic-structure systems

In the present work, the two-dimensional parametric acoustic-structure system consists of different mediums including liquid phase and solid phase. The liquid phase is filled with water, while the solid phase is composed of acoustic metasurface [38, 39] and linear elastic plate wall. The overall parametric acoustic-structure system of this study is shown in **Fig. 4**. The metasurface and the linear elastic wall below the interface $S^I$ of the incident wave are submerged in the exterior acoustic field $\Omega^E$. The whole system is divided into three computational domains and each domain obeys the Helmholtz PDEs under three specific conditions. First, the outer boundary obeys the plane wave radiation condition which surrounds the entire water medium domain in a quadrangle. Second, the water medium obeys the pressure acoustics condition. Third, the surface of the solid phase obeys the acoustic-structure coupling condition. Specifically, in the verification system, multi-physics phenomenon is realized by the coupling effect between acoustic metasurface and incident harmonic acoustic wave. The acoustic-structure coupling effect can be described by the generalized Snell's law [40] as

$$\theta_r = \arcsin\left(\frac{\lambda \partial \phi}{2\pi n \partial x}\right), \tag{18}$$

where $\theta_r$ denotes the reflection wave angle, $\lambda$ is the wavelength, $n$ represents the reflection indices, and $\partial \phi / \partial x$ represents the phase shifts along the $x$ direction. As an artificial layered metamaterial, the acoustic metasurface is artificially divided into multiple subunits. The metasurface is divided into 25 subunits and different physical parameters are assigned to each subunit. Each subunit is characterized by density and Young's modulus, resulting in a total of 50 physical parameters for acoustic metasurface. When the incident acoustic wave is emitted to the interface of the metasurface, it generates elastic deformation on the solid phase. The monitorable deformation of the solid causes phase shifts along the interface of different mediums. Therefore, the acoustic metasurface obeys the generalized Snell's law [40]. The phase shifts lead to deviation of the reflection acoustic wave and influence the solutions of scattered acoustic field, which is the consequence of the acoustic-structure coupling.

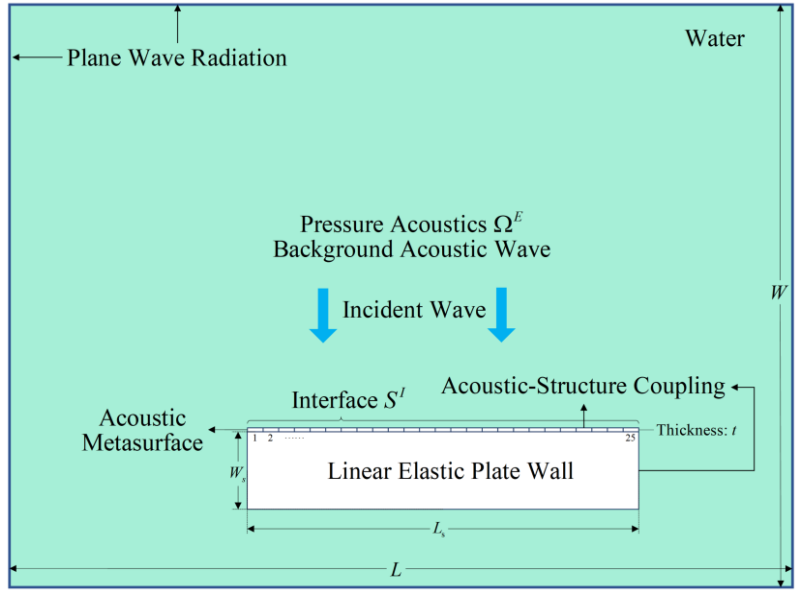

**Fig. 4** Overall parametric acoustic-structure system

## 4.1.2 Governing PDEs

The typical steady acoustic-structure system is composed of three computational domains and each part obeys corresponding governing PDEs. Solving the parametric acoustic-structure systems is predicting the solutions of the parametric Helmholtz equations across all computational domains. We apply harmonic background acoustic wave as the incident wave, which can be expressed as

$$p_\text{b} = p_0 e^{-ik_\text{eq}\left(\frac{\mathbf{X}\cdot e_\text{k}}{\|e_\text{k}\|}\right)}, \tag{19}$$

where the initial excitation pressure $p_0 = 1\text{pa}$, $e_\text{k}$ denotes the unit direction vector, X denotes the space coordinates. Specifically, $k_\text{eq} = (2\pi f)^2 / c$ denotes the exact value of the wave number under the specific velocity of the sound in water $c$ and frequency $f$ of the background acoustic wave. In the verification system, we set zero dipole domain source and zero monopole domain source. Therefore, we derive the governing PDEs for each domain as follows.

First, the water medium obeys the pressure acoustic condition, and the governing equation satisfies the wave equation form of the Helmholtz equation that can be expressed as

$$\nabla\left(-\frac{1}{\rho_\text{c}}(\nabla p_\text{t})\right) - \frac{k_\text{eq}^2 p_\text{t}}{\rho_\text{c}} = 0, \tag{20}$$

where $p_\text{t}$ is the total acoustic pressure, $\rho_\text{c}$ is the density of the water medium. Specifically, the total pressure $p_\text{t}$ is the sum of the scattered acoustic pressure $p_s$ and the background acoustic pressure $p_\text{b}$.

Second, the outer boundary obeys the plane wave radiation condition. Based on Givoli and Neta's reformulation of the Higdon conditions [41] for plane wave and Helmholtz equation. The outer boundary obeys the plane wave radiation and the governing equations can be expressed as

$$-\mathbf{n}\left(-\frac{1}{\rho_\text{c}}(\nabla p_\text{t})\right) + i\left(\frac{k_\text{eq}}{\rho_c} + \frac{\Delta_\|}{2k_\text{eq}\rho_c}\right)p_s = 0, \tag{21}$$

where $\Delta_\|$ at a given point on the boundary denotes the Laplace operator in the tangent plane at a particular point, $i$ is the imaginary unit, and $\mathbf{n}$ is the unit normal vector of the solid surface.

The boundary of the solid phase including the acoustic metasurface and the linear elastic plate wall obeys the acoustic-structure coupling condition. The steady state of the governing equation derived from harmonic wave and elastic condition of the Helmholtz equation can be expressed as

$$-\mathbf{n}\cdot\left(-\frac{1}{\rho_c}(\nabla p_t)\right) = -\mathbf{n}\cdot\left(-\omega^2 \bar{\mathbf{u}}\right), \tag{22}$$

where $\bar{\mathbf{u}}$ indicates the steady state displacement of the acoustic-structure boundaries which can be monitored directly, $\omega$ indicates angular velocity.

### 4.1.3 MPIPN for parametric acoustic-structure systems

We apply the MPIPN to solve the governing PDEs of the parametric acoustic-structure systems and study both explicit and implicit parametric conditions of the Helmholtz equations to determine complete solutions of the systems. The parametric conditions for the governing PDEs include the background acoustic frequency and physical parameters of the acoustic metasurface, where three physical quantities that govern the solutions are explicit and implicit in the PDEs of the system, respectively. Specifically, the background acoustic wave frequency is explicit, while the physical parameters of the acoustic metasurface are implicit.

The background acoustic wave frequency is directly existing in the PDEs. This can be expressed by substituting Eqs. (19) for Eqs (20)-(22) as

$$\nabla\left(-\frac{1}{\rho_c}\left(\nabla(p_s + p_0 e^{-i\frac{(2\pi f)^2}{c}\left(\frac{\mathbf{X}\cdot\mathbf{e}_k}{\|\mathbf{e}_k\|}\right)})\right)\right) - \frac{k_{eq}^2(p_s + p_0 e^{-i\frac{(2\pi f)^2}{c}\left(\frac{\mathbf{X}\cdot\mathbf{e}_k}{\|\mathbf{e}_k\|}\right)})}{\rho_c} = 0, \tag{23}$$

$$-\mathbf{n}\left(-\frac{1}{\rho_c}\left(\nabla(p_s + p_0 e^{-i\frac{(2\pi f)^2}{c}\left(\frac{\mathbf{X}\cdot\mathbf{e}_k}{\|\mathbf{e}_k\|}\right)})\right)\right) + i\left(\frac{k_{eq}}{\rho_c} + \frac{\Delta_\parallel}{2k_{eq}\rho_c}\right)p_s = 0, \tag{24}$$

$$-\mathbf{n}\cdot\left(-\frac{1}{\rho_c}\left(\nabla(\bar{p}_s + \bar{p}_0 e^{-i\frac{(2\pi f)^2}{c}\left(\frac{\mathbf{X}\cdot\mathbf{e}_k}{\|\mathbf{e}_k\|}\right)})\right)\right) = -\mathbf{n}\cdot\left(-\omega^2 \bar{\mathbf{u}}\right). \tag{25}$$

Therefore, the background acoustic wave frequency is an explicit quantity for the governing PDEs. Nevertheless, the physical parameters of the acoustic metasurface including density and Young's modulus are implicit quantities. Since the deviation effect on solutions of the system cannot be captured directly from the governing equations of the system, and the physical parameters do not exist as mutable variables in the governing

PDEs.In the validation cases, the MPIPN is used to solve parametric Helmholtz equations. The flowchart for the application of the MPIPN is shown in **Fig. 5**. $N_1$, $N_2$ and $N_3$ represent the number of points in point-cloud of the pressure acoustic domain, plane wave radiation boundary and acoustic-structure coupling boundary, respectively. $d_i$ and $d_q$ denote the dimensions of spatial coordinates and explicit quantities.

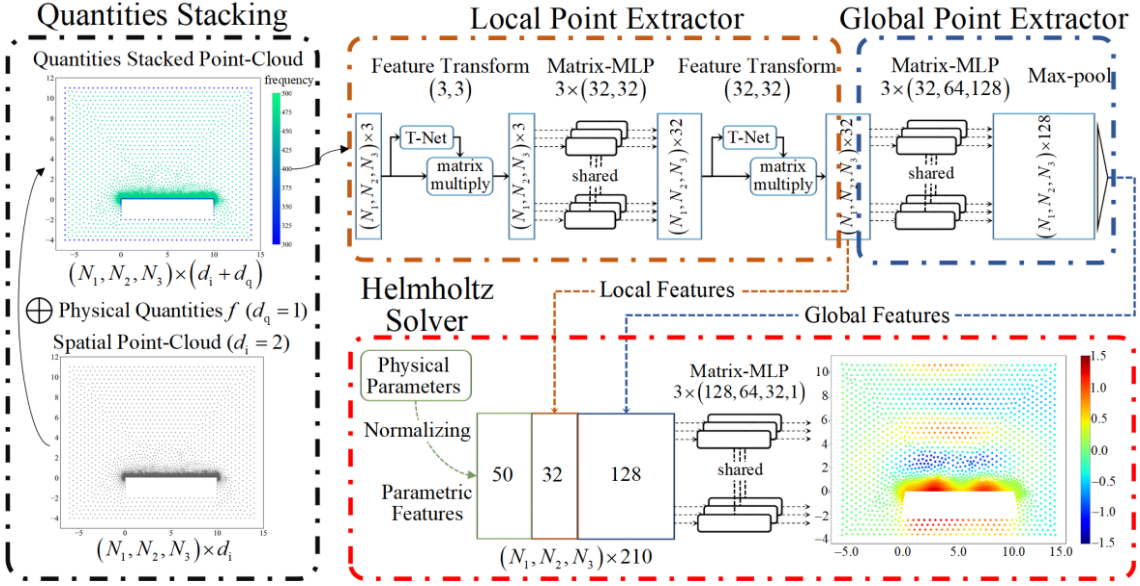

**Fig. 5** Application of the MPIPN for the parametric acoustic-structure systems

The Quantities Stacking module dimensionally stacks the explicit physical quantity, frequency of background acoustic wave $Q = \{q_i\}_{i=1}^{N} \in \mathbb{R}^{d_q}$ with initial spatial point-cloud $X = \{x_i\}_{i=1}^{N} \in \mathbb{R}^{d_i}$, where $d_q = 1$, and $d_i = 2$. The reconstructed point-cloud $(N_1, N_2, N_3) \times 3$ consists of two-dimensional spatial and frequency information on the three computational domains. The point-cloud is mapped by Local Point Extractor including Feature Transform ($\mathcal{F}(\cdot)$) and Matrix-MLP ($\mathcal{M}(\cdot)$) and outputs the local feature sequence $S_L$ $(N_1, N_2, N_3) \times 32$. The global feature sequence $S_G$ $(N_1, N_2, N_3) \times 128$ is obtained by mapping $S_L$ with the Global Point Extractor. As for the implicit physical quantities, physical parameters of the acoustic metasurface $Q^* = \{q_i^*\}_{i=1}^{N} \in \mathbb{R}^{d_q^*}$, we encode the combination of density and Young's modulus as a sequence through Z-Score normalization. Each quantity can be mathematically encoded as

$$\left(q_i^*\right)_{\text{new}} = \frac{q_i^* - \mu_{q_i^*}}{\sigma_{q_i^*}}, \text{ with } q_i^* \in \mathbb{R}^{d_q^*}, \tag{26}$$

where $d_q^*$ denotes dimensions of implicit quantities, $\mu_{q_i^*}$ indicates the mean of the implicit quantity, and $\sigma_{q_i^*}$ indicates the standard deviation of the implicit quantity. By using the Z-Score normalization as an encoder, the implicit quantities can be scaled to the same level as the parametric sequence $S_P$ $(N_1, N_2, N_3) \times 50$. The concatenating solving criteria sequence $(N_1, N_2, N_3) \times 210$ is mapped to output the solutions as

$$Y = \mathcal{M}\big((N_1, N_2, N_3) \times 210\big), \text{ with } Y \in \mathbb{R}^1 \tag{27}$$

where $\mathcal{M}(\cdot)$ indicates the Matrix-MLP $3 \times (128, 64, 32, 1)$ that maps the criteria sequence $(N_1, N_2, N_3) \times 210$ to solutions $Y$ $(N_1, N_2, N_3) \times 1$. All mapping networks including $\mathcal{F}(\cdot)$ and $\mathcal{M}(\cdot)$ necessitate being trained on $\Gamma$ with priori knowledge of $X = \{x_i\}_{i=1}^{N} \in \mathbb{R}^{d_i}$, $Q = \{q_i\}_{i=1}^{N} \in \mathbb{R}^{d_q}$, $Q^* = \{q_i^*\}_{i=1}^{N} \in \mathbb{R}^{d_q^*}$ and the Helmholtz equations under constraints of physics-informed loss function and observation loss function.

For the three computational domains in the acoustic-structure system, physics-informed loss functions are derived for each domain. We define: $\mathcal{L}_{\text{pad}}$ as the loss function of pressure acoustic domain governed by Eqs. (20), $\mathcal{L}_{\text{pwr-r}}$ and $\mathcal{L}_{\text{pwr-i}}$ as the real and imaginary part, respectively for the loss function of the plane wave radiation boundary governed by Eqs. (21), $\mathcal{L}_{\text{asc}}$ as the loss function of acoustic-structure coupling boundary governed by Eqs. (22), and $\mathcal{L}_{\text{OBS}}$ as the residual of the priori observation solutions and the predicted values on those selected discrete points. The above loss function can be expressed as

$$\mathcal{L}_{\text{pad}} = \frac{1}{N_1} \sum_{i=1}^{N_1} \left\| \left(\frac{\partial^2 \tilde{p}_s}{\partial x^2} + \frac{\partial^2 \tilde{p}_s}{\partial y^2}\right) + \left(\frac{\partial^2 p_b}{\partial x^2} + \frac{\partial^2 p_b}{\partial y^2}\right) + k_{\text{eq}}^2 \left(\tilde{p}_s + p_b\right) \right\|_1, \tag{28}$$

$$\mathcal{L}_{\text{pwr-r}} = \frac{1}{N_2} \sum_{i=1}^{N_2} \left\| -\mathbf{n} \cdot \left(\frac{\partial \tilde{p}_s}{\partial x} \boldsymbol{x} + \frac{\partial \tilde{p}_s}{\partial y} \boldsymbol{y} + \frac{\partial p_b}{\partial x} \boldsymbol{x} + \frac{\partial p_b}{\partial y} \boldsymbol{y}\right) \right\|_1, \tag{29}$$

$$\mathcal{L}_{\text{pwr-i}} = \frac{1}{N_2} \sum_{i=1}^{N_2} \left\| k_{\text{eq}} \cdot \tilde{p}_s + \frac{1}{2k_{\text{eq}}} \left(\frac{\partial^2 \tilde{p}_s}{\partial x^2} \boldsymbol{x} + \frac{\partial^2 \tilde{p}_s}{\partial y^2} \boldsymbol{y}\right)_{\text{tan}} \right\|_1, \tag{30}$$

$$\mathcal{L}_{\text{asc}} = \frac{1}{N_3} \sum_{i=1}^{N_3} \left\| \mathbf{n} \cdot \left( \left( \frac{\partial^2 \bar{p}_s}{\partial x^2} x + \frac{\partial^2 \bar{p}_s}{\partial y^2} y \right) + \left( \frac{\partial^2 p_b}{\partial x^2} x + \frac{\partial^2 p_b}{\partial y^2} y \right) \right) - \mathbf{n} \cdot \left( \omega^2 \mathbf{u} \right) \right\|_1, \tag{31}$$

$$\mathcal{L}_{\text{OBS}} = \frac{1}{N_4} \sum_{i=1}^{N_4} \left\| \bar{p}_s - p_s \right\|_2, \tag{32}$$

where $N_4$ refers to the number of observation solutions in the system, $\tilde{p}_s$ denotes predicted solutions and $p_s$ denotes ground true solutions. The observation solutions are randomly selected to distribute and kept constant in the system while training. This avoids extra knowledge induced by manual design of the position of observation points (sensors when implementing experiments), thus ensuring robustness of the MPIPN for general acoustic-structure systems. The reason why observation solutions are needed is mathematically deduced from the degenerate solutions of PDEs. For Eqs. (20) of pressure acoustic domain, one degenerate solution can be obtained that $p_t = 0$, and the scattered acoustic pressure $p_s = -p_b$ for all points in the domain, which is not the solution we want for the system. For Eqs. (21) of the plane wave radiation boundary, one degenerate solution of the real part of the equation is $p_s = -p_b$, whereas one degenerate solution of the imaginary part is $p_s = 0$. Since $p_b = 1\text{pa}$, the neural networks tend to obtain a balanced value between 0 and 1 to minimize the residual of the loss function for governing equations. These will affect the solutions of the acoustic-structure coupling boundary.

In the numerical experiments, we utilized an optimizer that combines RAdam (Rectified Adam) [42] and LookAhead [43], RAdam-LookAhead to solve the optimization problem of parameters updating in models. To compare the performance of MPIPN fairly, the framework was trained three times in the same environment and the average results are recorded. We obtained the predicted solutions $\tilde{p}_s$ of the Helmholtz equations on three computational domains. Relative domain errors of $L_2$ norm (RDE) and absolute pointwise errors of $L_1$ norm (APE) are utilized as evaluation metrics for each computational domain, which are given by

$$\text{RDE} = \sum_{i=1}^{N} \left\| \tilde{p}_{s_i} - p_{s_i} \right\|_2 / \sum_{i=1}^{N} \left\| p_{s_i} \right\|_2, \tag{33}$$

$$\text{APE} = \left| \tilde{p}_{s_i} - p_{s_i} \right|_1, \tag{34}$$

where $N$ indicates the number of points for the computational domain, respectively. RDE provides the global accuracy of predicted solutions, and APE intuitively reflects the deviation of the predicted solutions.

## 4.2 Constant parametric conditions

The MPIPN is tested to solve the Helmholtz equations on a series of constant parametric conditions for the systems. In the acoustic field $\Omega^E$ of the system, the length of the acoustic domain $L = 20\text{m}$, the width of the domain $W = 15\text{m}$, the length of the linear elastic plate wall $L_s = 10\text{m}$ and the width of the linear elastic plate wall $W_s = 2\text{m}$. The acoustic metasurface is in an extremely thin shape, whose thickness is less than the acoustic wavelength. Therefore, the thickness of the metasurface $t = 0.08\text{m}$. After spatial discretization, the oriented PDEs are as follows

$$\partial_x^2 p_\text{s} + \partial_y^2 p_\text{s} + \partial_x^2 p_\text{b} + \partial_y^2 p_\text{b} + k_\text{eq}^2 (p_\text{s} + p_\text{b}) = 0, \quad \begin{array}{l} x \in (-10,-5] \cup [5,10), y \in (-4,11), \\ x \in [-5,5], y \in (-4,-2] \cup [0.08,11), \end{array} \quad (35)$$

$$\begin{cases} \mathbf{n} \cdot (\partial_x p_\text{s} \cdot \mathbf{x} + \partial_y p_\text{s} \cdot \mathbf{y} + \partial_x p_\text{b} \cdot \mathbf{x} + \partial_y p_\text{b} \cdot \mathbf{y}) = 0, \\ k_\text{eq} \cdot p_\text{s} + \dfrac{1}{2k_\text{eq}} (\partial_x^2 p_\text{s} \cdot \mathbf{x} + \partial_y^2 p_\text{s} \cdot \mathbf{y})_\text{tan} = 0, \end{cases} \quad x = -10, x = 10, y = -4, y = 11, \quad (36)$$

$$\mathbf{n} \cdot (\partial_x^2 p_\text{s} \cdot \mathbf{x} + \partial_y^2 p_\text{s} \cdot \mathbf{y} + \partial_x^2 p_\text{b} \cdot \mathbf{x} + \partial_y^2 p_\text{b} \cdot \mathbf{y}) = \mathbf{n} \cdot \omega^2 \mathbf{u}, \quad \begin{array}{l} x \in [-5,5], y = -2 \text{ or } 0.08, \\ x = -5 \text{ or } 5, y \in [-2, 0.08]. \end{array} \quad (37)$$

In the following case 1 and case 2, there are $N_1 = 1377$ points in the pressure acoustic domain, $N_2 = 88$ points on the plane wave radiation boundary, $N_3 = 158$ points on the acoustic-structure coupling boundary. Overall, a total of 1623 oriented discrete points necessitate to be solved in case 1 and case 2. As for observation solutions, 50 discrete points are randomly selected to provide the ground truth as priori physical constraints. Therefore, for loss function Eqs. (32), the number of observation solutions is $N_4 = 50$. Among the selected priori solutions, there are 4 solutions for the plane wave radiation, 30 solutions for the pressure acoustic domain and 16 solutions for the acoustic-structure coupling boundary. The extracted point-cloud of the computational domain and the observation discrete points are illustrated in **Fig. 6**.

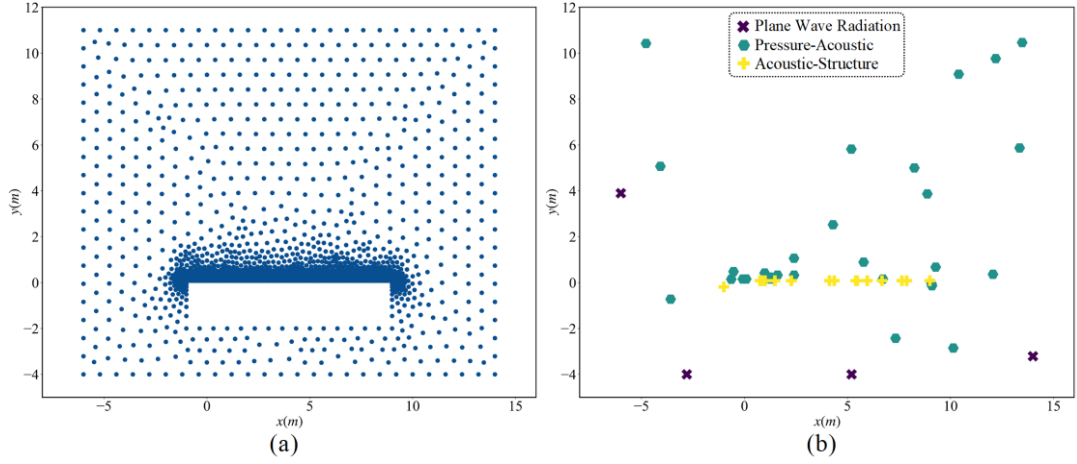

**Fig. 6** Discrete point-cloud of case 1 and case 2: (a) overall oriented discrete points, (b) randomly selected observation points

### 4.2.1 Case 1: Constant background acoustic wave frequency

In case 1, the MPIPN is tested to validate its ability to extract explicit physical quantity, which is the background acoustic wave frequency in the system. We select three background acoustic wave frequencies as a constant variable, which are $300\,\mathrm{Hz}$, $350\,\mathrm{Hz}$, $400\,\mathrm{Hz}$, $450\,\mathrm{Hz}$, $500\,\mathrm{Hz}$. For each constant frequency, we set 1000 changeable physical parameters. The density of each subunit is randomly selected from the interval ranging from $1/3$ to 2 times the water density, where the water density is $1000\,\mathrm{kg/m^3}$. The Young's modulus of each subunit is randomly selected from the interval ranging from $1/3$ to 5 times the water modulus, where the Young's modulus of water is $2.25\times10^6\,\mathrm{Pa}$. For the linear elastic wall, Poisson's ratio is set to be 0.34 and the density is $4500\,\mathrm{kg/m^3}$, with Young's modulus selected to be $1.08\times10^5\,\mathrm{MPa}$. The box plot of RDE on four frequencies with 1000 combinations of parameters is shown in **Fig. 7**. The maximum error is less than $2\times10^{-2}$, and the average error is less than $1\times10^{-2}$ With the frequency increasing, the average RDE and dispersion of errors are in reasonable interval. This indicates that, with constant explicit physical quantities, the MPIPN is able to identify the implicit quantities.

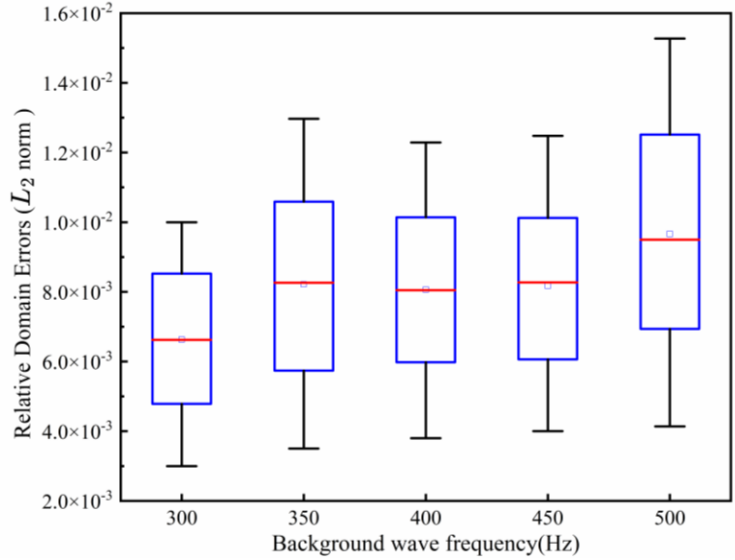

**Fig. 7** Box plot of Relative domain errors on testing background wave frequencies

### 4.2.2 Case 2: Constant physical parameters

In case 2, the MPIPN is tested to validate its ability to encode and fuse the implicit physical quantities to solve the related parametric PDEs. In the acoustic-structure system, physical parameters including density and Young's modulus are implicit quantities. We select three combinations of physical parameters as constant variables, which are densities of 1, 1.5, 2 times the water density and Young's modulus of 1, 3, 5 times the water modulus. The parameters of elastic plate wall are the same as it in case 1. For each combination, the background wave frequency uniformly sampled 1000 frequencies from 300 Hz to 500 Hz. We present one combination of physical parameters of the metasurface computed under frequencies of 300 Hz, 400 Hz, 500 Hz in **Fig. 8**. The mean APE of 300 Hz, 400 Hz, 500 Hz are 0.017, 0.008 and 0.013, respectively. It can be noted that although various frequencies cause great change in the distribution of solutions, the absolute errors are at reasonably low level. The quantitative average RDE of one randomly selected combination of physical parameters on the testing frequencies is illustrated in **Table. 1**. The proposed method successfully identifies the background wave frequency and maintains considerable accuracy. The MPIPN is able to capture the changes in frequency as explicitly existing in the governing PDEs with constant implicit quantities.

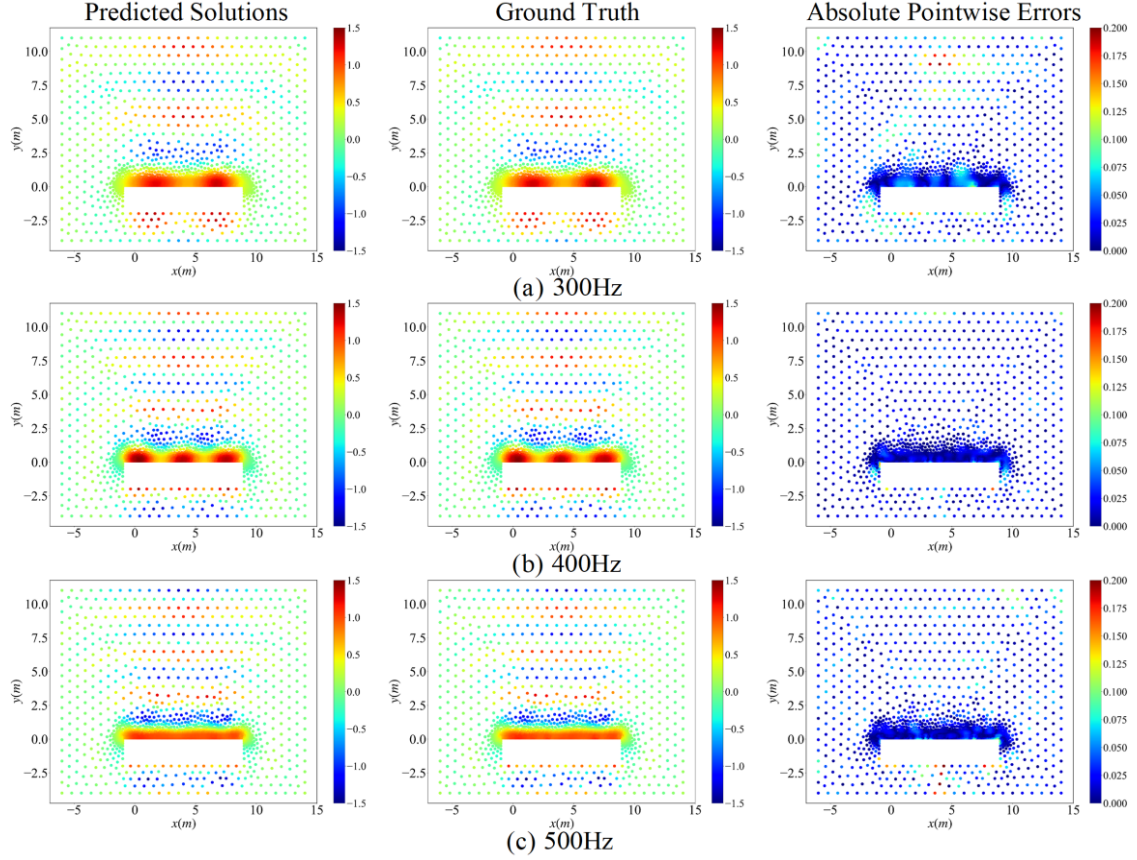

**Fig. 8** Predicted solutions compared with ground truth in absolute pointwise errors level

**Table. 1** RDE of one combination of physical parameters on three frequencies

| Frequency (Hz) | Average $\|\tilde{p}_s - p_s\|_2 / \|p_s\|_2$ on $\Lambda$ | | |
|---|---|---|---|
| | Pressure Acoustics | Plane Wave Radiation | Acoustic-Structure Coupling |
| 300 | 6.4285E-3 | 1.3242E-2 | 7.8252E-3 |
| 400 | 4.1452E-3 | 9.1213E-3 | 6.3256E-3 |
| 500 | 4.9424E-3 | 1.1623E-2 | 7.4367E-3 |
| **Average** | 5.1720E-3 | 1.1329E-2 | 7.1958E-3 |
| **Variance** | 1.3429E-6 | 4.3099E-6 | 6.0571E-7 |

### 4.3 Case 3: Changeable parametric conditions

In case 3, the MPIPN is tested to solve the Helmholtz equations with changeable combinations of background frequency and physical parameters of the acoustic metasurface. Both the explicit and implicit conditions are changeable and form 1000 combinations of parametric conditions as training datasets $\Gamma$ and 200 unseen combinations as testing datasets $\Lambda$. Specifically, the overall 1200 combinations are composed of 12 groups of physical parameters and 100 frequencies. All quantities are randomly selected from the interval mentioned in the constant condition cases. In case 3, we adjust the position of the solid phase and create new oriented PDEs as follows

$$\partial_x^2 p_s + \partial_y^2 p_s + \partial_x^2 p_b + \partial_y^2 p_b + k_{eq}^2 (p_s + p_b) = 0, \quad \begin{array}{l} x \in (-6,0] \cup [10,14), y \in (-4,11), \\ x \in [0,10], y \in (-4,-2] \cup [0.08,11), \end{array} \quad (38)$$

$$\begin{cases} \mathbf{n} \cdot (\partial_x p_s \cdot \mathbf{x} + \partial_y p_s \cdot \mathbf{y} + \partial_x p_b \cdot \mathbf{x} + \partial_y p_b \cdot \mathbf{y}) = 0, \\ k_{eq} \cdot p_s + \dfrac{1}{2k_{eq}} (\partial_x^2 p_s \cdot \mathbf{x} + \partial_y^2 p_s \cdot \mathbf{y})_{\tan} = 0, \end{cases} \quad x = -6, x = 14, y = -4, y = 11, \quad (39)$$

$$\mathbf{n} \cdot (\partial_x^2 p_s \cdot \mathbf{x} + \partial_y^2 p_s \cdot \mathbf{y} + \partial_x^2 p_b \cdot \mathbf{x} + \partial_y^2 p_b \cdot \mathbf{y}) = \mathbf{n} \cdot \omega^2 \mathbf{u}, \quad \begin{array}{l} x \in [0,10], y = -2 \text{ or } 0.08, \\ x = 0 \text{ or } 10, y \in [-2, 0.08]. \end{array} \quad (40)$$

There are $N_1 = 4928$ points in the pressure acoustic domain, $N_2 = 140$ points on the plane wave radiation boundary, $N_3 = 554$ points on the acoustic-structure coupling boundary. Overall, a total of 5622 oriented discrete points necessitate to be solved in case 3. As for observation solutions, 150 discrete points are randomly selected to provide the ground truth as priori physical constraints. For loss function Eqs. (32), the number of observation solutions is $N_4 = 150$. Among the priori solutions, there are 10 solutions for the plane wave radiation, 100 solutions for the pressure acoustic domain and 40 solutions for the acoustic-structure coupling boundary. We propose the extracted point-cloud of the computational domain and the observation discrete points in **Fig. 9**.

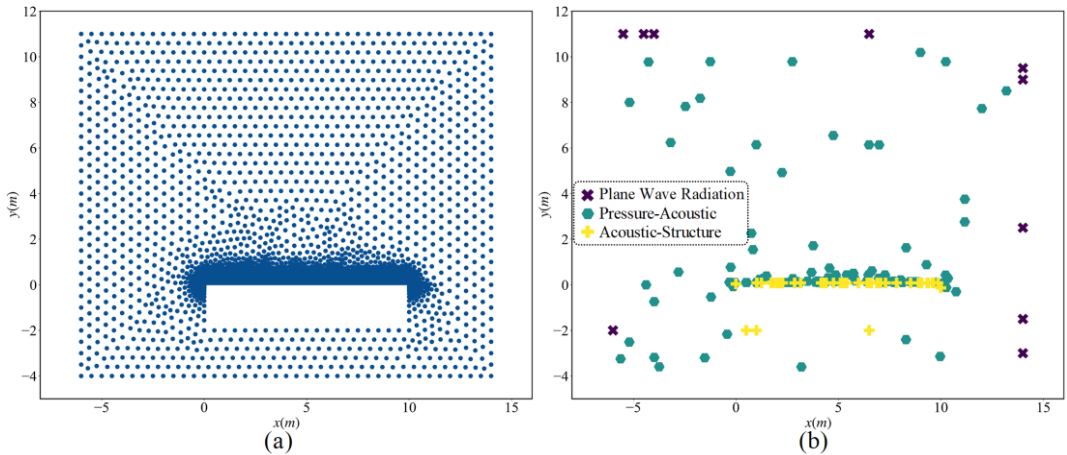

**Fig. 9** Discrete point-cloud of case 3: (a) overall oriented discrete points, (b) randomly selected observation points

### 4.3.1 General Results

Among the testing dataset $\Lambda$, we randomly selected an example of predicted solutions and ground truth. The results are compared and the APE are presented in **Fig. 10**. It can be observed that predicted solutions are consistent with the ground truth distribution, proving that the proposed framework captures the mapping functions to features of various scales and features of different computational domains. Solutions near the acoustic-structure boundary are the most significant parts of the system, since the coupling conditions are complex and the exact value of the solutions are in a strong correlation with the values of the acoustic pressure field. Besides, the predicted solutions near the coupling boundary are highly accurate. The error analysis of the solutions compared with ground true values is shown in **Table. 2**. The framework obtained predicted solutions with an average RDE of $1.52\%$ on the pressure acoustic domain, $2.03\%$ on the plane wave radiation boundary and $1.64\%$ on the acoustic-structure coupling boundary. The maximum RDE for unseen parametric conditions in $\Lambda$ are less than $5\%$. **Fig. 11** shows the average absolute errors of $\mathcal{L}_{\text{PDE}}$ on $\Lambda$. The errors distribution is at reasonable level of $10^{-3}$. The intuitive distribution of relative errors on each domain and the probability density function (PDF) of RDE for all domains are shown in **Fig. 12**. The distributions of errors on three computational domains are close to the normal distributions. Fewer outliers in the error distribution exist in the histogram, which is an acceptable margin of error for solving the oriented equations.

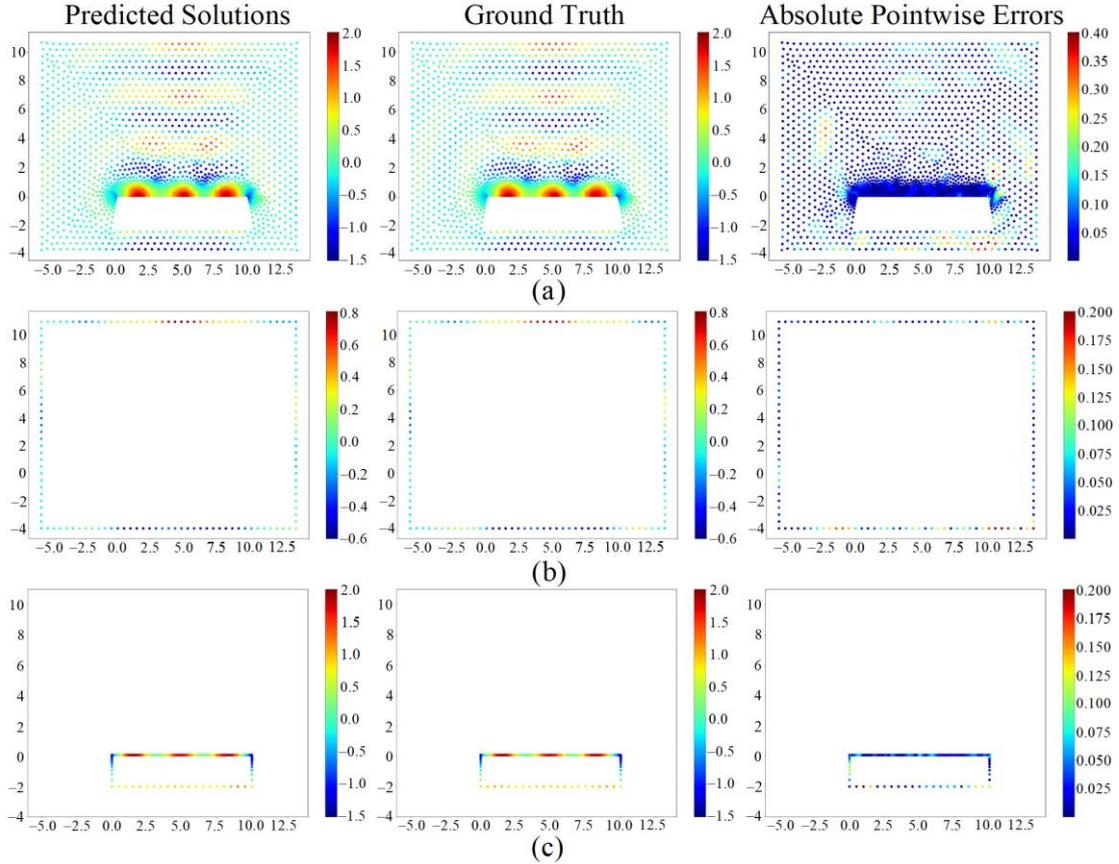

**Fig. 10** Randomly selected predicted results: (a) Pressure acoustic domain, (b) Plane wave radiation boundary, (c) Acoustic-structure coupling boundary

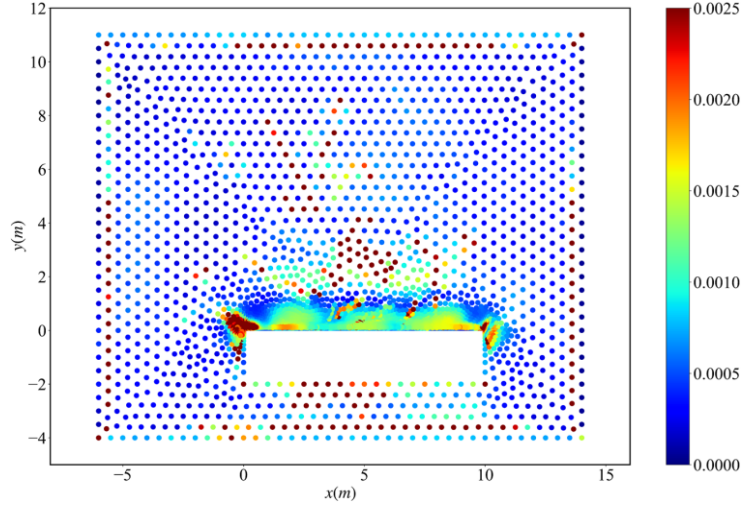

**Fig. 11** Average absolute errors of $\mathcal{L}_{\text{PDE}}$ on testing datasets $\Lambda$

**Table. 2** Comparison results of different computational domains on $\Lambda$

| Computational Domains | Average $\|\tilde{p}_s - p_s\|_2 / \|p_s\|_2$ | Maximum $\|\tilde{p}_s - p_s\|_2 / \|p_s\|_2$ | Minimum $\|\tilde{p}_s - p_s\|_2 / \|p_s\|_2$ |
|---|---|---|---|
| Pressure Acoustics | 1.5171E-2 | 4.9791E-2 | 9.1626E-3 |
| Plane Wave Radiation | 2.0311E-2 | 3.7654E-2 | 1.1523E-2 |
| Acoustic-Structure Coupling | 1.6438E-2 | 8.7362E-2 | 7.4959E-3 |

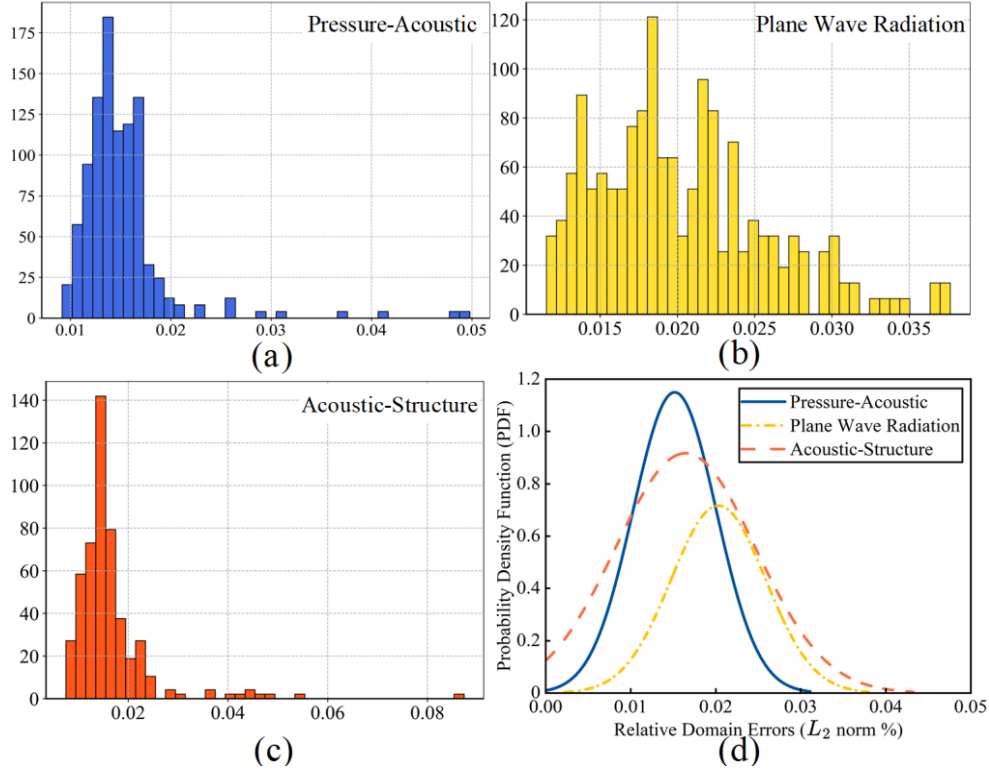

**Fig. 12** Relative domain errors: (a) Pressure acoustic domain, (b) Plane wave radiation boundary (c) Acoustic-structure boundary, (d) PDF of relative domain errors

## 4.3.2 Errors clustering phenomenon with training stages

A noteworthy phenomenon in **Fig. 10** is that, those points whose APE are relatively high are clustering in the domain. The reason is that the Helmholtz equations under the pressure acoustic condition constrain the gradient information of the solutions for the point-cloud. For a cluster of points, the gradients inside satisfy the governing equations but values of points are not exactly right. To illustrate this, we take the pressure acoustic domain as example since it contains the majority of arbitrary discrete points in acoustic-structure systems. **Fig. 13** shows the predicted solutions obtained by MPIPN after training on datasets Γ at the beginning, middle and end stages of the training process, which are 200, 800, and 1800 epochs respectively. At the beginning of the training, the predicted solutions are inaccurate. The number of clusters of points with high errors and the distribution area are large. When trained to the middle stage, clusters with high errors begin to shrink, and the general trend of the solutions is right. After being trained to convergence, the predicted solutions fit in well with the ground truth and few clusters of high errors exist, meaning that the governing equations of the domain are fully satisfied. Noticing that the clusters of high errors first fade away at areas with high point-cloud density (near the metasurface), and then is the area with low point-cloud density. This can be explained by the convergence and divergence of the meshes during computation: The high density of meshes makes the derivative computation between points more accurate. In this sense, although the high density of point-cloud makes the amount of calculation increase and harder for the training of the neural network, it is more suitable for the fitness of the governing equations. This indicates that the well-trained MPIPN is capable of capturing the gradient information of point-cloud in clusters.

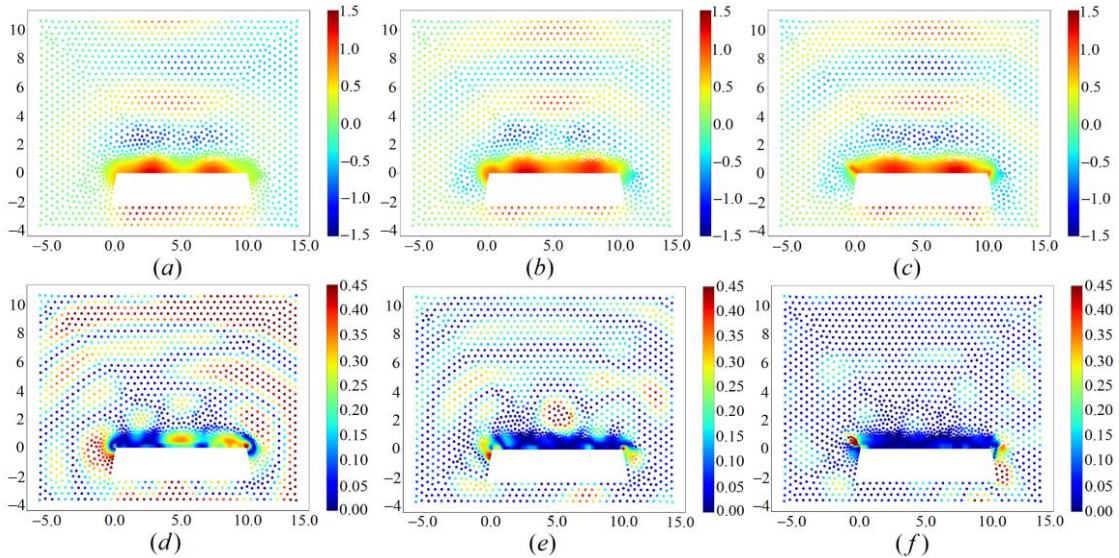

**Fig. 13** Comparison between predicted solutions and ground truth of pressure acoustic domain: (a) 200 epochs, (b) 800 epochs, (c) 1800 epochs. (d), (e), (f) correspond to the APE of (a), (b), (c) respectively

### 4.3.3 Explicit background acoustic wave frequency conditions

As the explicit parametric condition for the acoustic-structure system, frequency of the background acoustic wave determines the background wave pressure and directly influences the solutions of scattered acoustic pressure on three domains, which can be inferred from Eqs. (23), (24) and (25). Different domains computed in the same frequency form a complete scattered acoustic field under certain frequency condition. Here, we show the predicted solutions on $300\,\text{Hz}$, $350\,\text{Hz}$, $400\,\text{Hz}$, and $500\,\text{Hz}$ across the sampling interval with the same physical parameters of the metasurface in **Fig. 14**. It can be noted from the comparison that the frequency generates significant variance of the acoustic pressure field. As the frequency increases from $300\,\text{Hz}$ to $500\,\text{Hz}$, the gradient of the pressure distribution of the scattered field changes more sharply. The average RDE for multi-frequency condition are shown in **Table. 3**. The average RDE are less than $2.0\%$ on all domains. The values of errors tend to be smaller around $440\,\text{Hz}$ than those on other frequencies. This is because the priori knowledge is densely distributed at about $400\,\text{Hz}$ and the distribution of frequency sampling points is smoother around 400hz, making neural networks have a greater weight for updating the predicted solutions at this frequency interval to reduce the loss function rapidly and the correlation between frequency and spatial coordinates are prominently expressed. It can be observed that there are no outliers for four frequencies with maximum variance of RDE less than $4\times10^{-5}$. This indicates the proposed framework is stable across frequencies. Therefore, the frequency parametric condition for the acoustic-structure system can be distinguished by MPIPN. The framework can capture the correlation between the explicit physical quantities and the initial spatial information, and accurately solve the Helmholtz equations concerning varying explicit physical quantities.

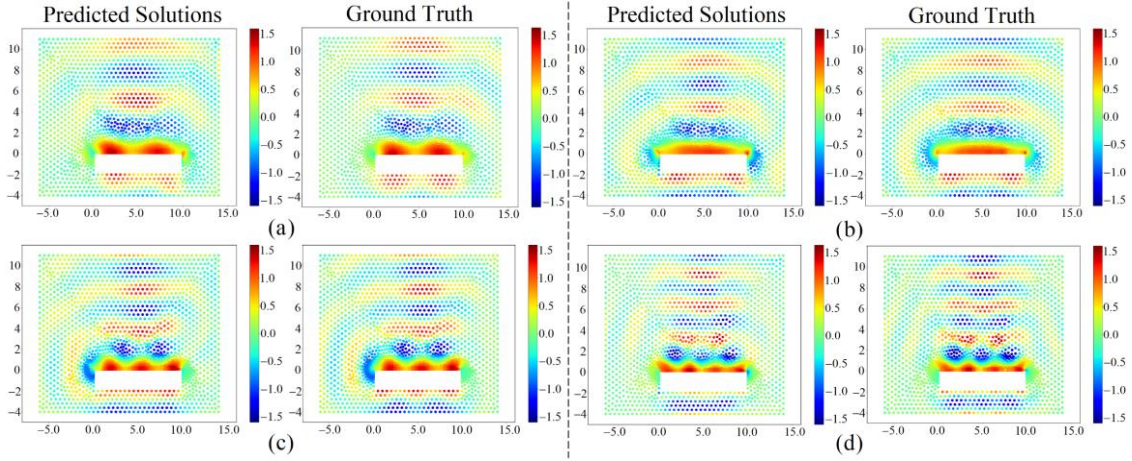

**Fig. 14** Predicted solutions and corresponding ground truth on multi-frequency: (a) 300 Hz, (b) 350 Hz, (c) 400 Hz, (d) 500 Hz

**Table. 3** Average relative domain errors for changeable frequencies

| Frequency (Hz) | Average $\|\tilde{p}_s - p_s\|_2 / \|p_s\|_2$ on $\Lambda$ | | |
| --- | --- | --- | --- |
| | Pressure Acoustics | Plane Wave Radiation | Acoustic-Structure Coupling |
| 300 | 1.4422E-2 | 1.9922E-2 | 1.4299E-2 |
| 320 | 1.6028E-2 | 2.3603E-2 | 1.4846E-2 |
| 350 | 1.4910E-2 | 2.1561E-2 | 2.1490E-2 |
| 380 | 1.3544E-2 | 2.3812E-2 | 1.1810E-2 |
| 400 | 1.3205E-2 | 1.8333E-2 | 1.4434E-2 |
| 420 | 1.2765E-2 | 1.8414E-2 | 1.6156E-2 |
| 450 | 1.2097E-2 | 1.1545E-2 | 9.7292E-3 |
| 480 | 1.5844E-2 | 1.6679E-2 | 2.3760E-2 |
| 500 | 1.8186E-2 | 1.7468E-2 | 2.8021E-2 |
| **Average** | **1.4556E-2** | **1.9037E-2** | **1.7172E-2** |
| **Variance** | **3.2505E-6** | **1.2871E-5** | **3.1789E-5** |

### 4.3.4 Implicit physics parameters conditions

The 12 sets of physical parameters of the metasurface are randomly selected on the proposed interval. A variety of physical parameters causes phase shifts to generate deviated angles on the reflection acoustic wave physically. Mathematically, the variance of the physical properties changes the acoustic-structure coupling condition by affecting the displacement field. Therefore, variances of physical properties of the metasurface have the greatest influence on the solutions on acoustic-structure coupling boundary and indirectly affect other domains. We present predicted solutions of 6 groups of physical parameters out of 12 groups in the $\Gamma$ under the frequency of 450 Hz and corresponding ground truths in **Fig. 15**. In comparison with multi-frequency condition, it can be observed that the proposed framework ensures that the distribution trend of solutions with the same frequency is consistent when the physical parameter vector changes and capture the variations near the coupling boundary and points near the boundary. Detailed values of the RDE are shown in **Table. 4**. The average RDE on all physical parameters are less than 1.8% and the maximum variance of RDE is less than $1\times10^{-5}$, proving that the proposed framework is stable across the 12 sets of physical parameters. This implicit parametric condition of the Helmholtz equations can be identified by the framework. In fact, by fusing combination of density and Young's modulus directly with features of point-cloud, physical parameters that lack a direct mathematical correlation with PDEs and do not have a functional relationship with explicit physical quantities within the system can be directly utilized as criteria for the solutions. The MPIPN successfully constructs a mapping relationship between implicit physical quantities and corresponding solutions of the acoustic-structure systems.

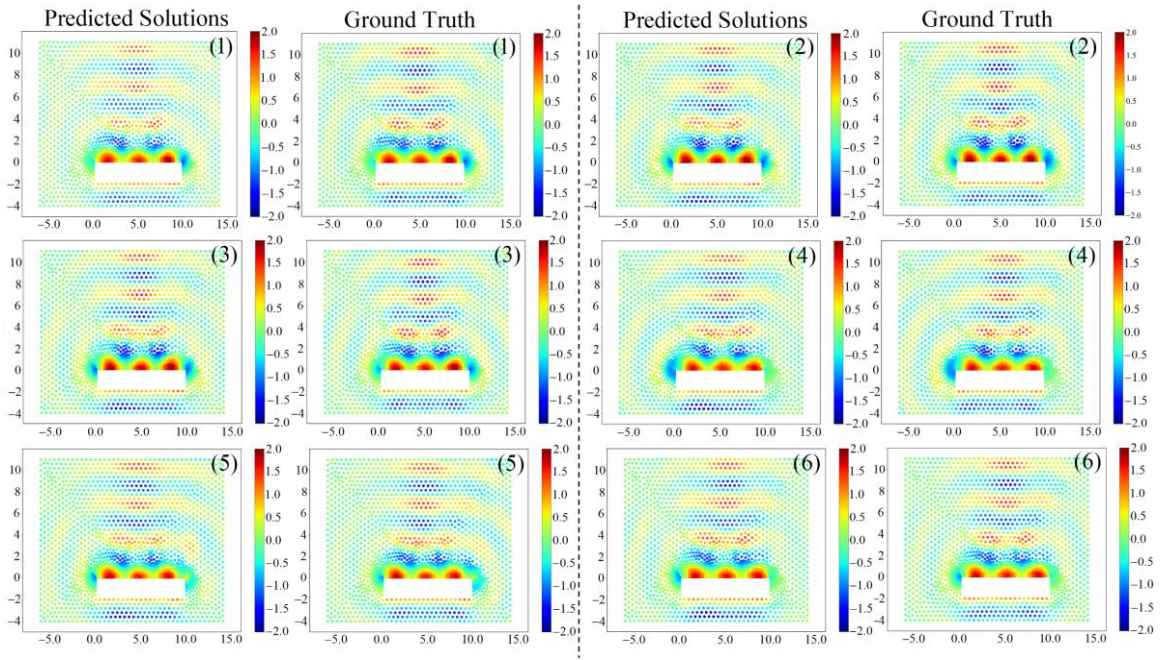

**Fig. 15** Predicted solutions and corresponding ground truth on multi-physical parameters: (1) to (6) denote different combinations of physical parameters

**Table. 4** Average relative domain errors for changeable physical parameters

| Combination of physical parameters | Average $\|\tilde{p}_s - p_s\|_2 / \|p_s\|_2$ on $\Lambda$ | | |
|---|---|---|---|
| | Pressure Acoustics | Plane Wave Radiation | Acoustic-Structure Coupling |
| 1 | 1.3632E-2 | 1.6578E-2 | 1.4999E-2 |
| 2 | 1.3755E-2 | 1.6931E-2 | 1.0496E-2 |
| 3 | 1.3371E-2 | 1.6282E-2 | 1.1630E-2 |
| 4 | 1.3750E-2 | 1.6452E-2 | 1.4161E-2 |
| 5 | 1.6537E-2 | 2.0853E-2 | 1.5661E-2 |
| 6 | 1.2849E-2 | 1.4523E-2 | 1.0548E-2 |
| 7 | 1.5097E-2 | 1.4546E-2 | 1.2729E-2 |
| 8 | 1.3815E-2 | 1.6735E-2 | 1.3718E-2 |
| 9 | 1.5466E-2 | 2.2076E-2 | 1.7139E-2 |
| 10 | 1.4222E-2 | 2.0873E-2 | 1.4908E-2 |
| 11 | 1.3839E-2 | 1.8527E-2 | 1.6835E-2 |
| 12 | 1.5547E-2 | 1.5640E-2 | 1.9253E-2 |
| **Average** | **1.4323E-2** | **1.7501E-2** | **1.4340E-2** |
| **Variance** | **1.0848E-6** | **5.8338E-6** | **6.6846E-6** |

## 4.4 Ablation experiment

We train the MPIPN with only observation solutions as data-driven regime to validate the effectiveness of physics-informed regime as the ablation experiment. At the same time, the comparison of data-driven and physics-informed methods is completed. It is used to prove that the proposed model fully utilizes the mathematical constitutive information of PDE to obtain the solutions. Here we present the results on the pressure acoustic domain in **Fig. 16**. The randomly selected observation solutions in the domain mainly crowd on the upper of the metasurface, making this area have relatively high accuracy. However, for areas that have a low density of observation solutions, the APE are unacceptably high. Quantitative comparison results are tabulated in **Table. 5**. In comparison with training by $\mathcal{L}_{OBS}$, the MPIPN that is trained with $\mathcal{L}_{PDE}$ and $\mathcal{L}_{OBS}$ successfully captures the regulations of PDEs and the values of residual remain at a low level. Therefore, this verifies the validity of the physical information in solving the equation and also proves that the MPIPN is capable of identifying and transforming characteristics of computational conditions and parametric conditions as criteria for solving the Helmholtz equations. The MPIPN ensures that the constitutive mathematical information of PDE can be used to solve pointwise solutions of the entire computational domain with high accuracy while providing a priori solutions for less than 3％ of the total number of points.

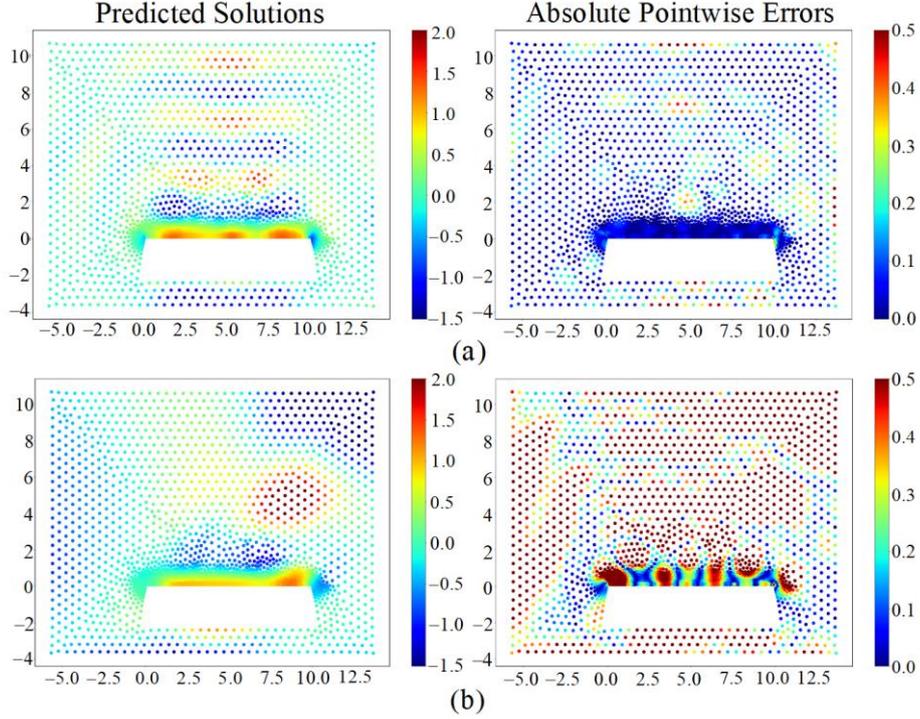

**Fig. 16** Comparison between: (a) training with physics-informed regime, (b) training with observation solutions as data-driven regime

**Table. 5** Comparison results of physics-informed and data-driven regimes

| Computational Domains | | Pressure Acoustics | Plane Wave Radiation | Acoustic-Structure Coupling |
|---|---|---|---|---|
| Average $\|\tilde{p}_s - p_s\|_2 / \|p_s\|_2$ | Physics-informed | **1.5171E-2** | **2.0311E-2** | **1.6438E-2** |
| | Data-driven | 9.4462E-2 | 8.2960E-1 | 6.1231E-2 |
| Maximum $\|\tilde{p}_s - p_s\|_2 / \|p_s\|_2$ | Physics-informed | **3.9791E-2** | **3.7654E-2** | **7.7362E-2** |
| | Data-driven | 1.9638E-1 | 8.1457E-1 | 1.5955E-1 |
| Minimum $\|\tilde{p}_s - p_s\|_2 / \|p_s\|_2$ | Physics-informed | **9.1626E-3** | **1.1523E-2** | **7.4959E-3** |
| | Data-driven | 3.1438E-2 | 4.0934E-2 | 1.7630E-2 |

## 5 Conclusions

In this paper, we proposed Multi Physics-informed PointNet (MPIPN) to solve the parametric acoustic-structure systems which are classic and vital multi-physics systems. Both explicit and implicit parametric conditions that govern the acoustic-structure systems can be solved. The MPIPN extracts the spatial geometric information and embeds parametric physical quantities by using a developed point-cloud architecture. The framework fuses the explicit physical quantities with the spatial point-cloud by dimensionally Quantities Stacking module. The correlation between physical quantities can be captured and encoded to sequences by Local Point Extractor and Global Point

Extractor. Meanwhile, for implicit physical quantities that are not directly expressible to the governing PDEs, the MPIPN embeds these quantities by manually encoding method automatically or statistically. The proposed framework is trained by adaptive physics-informed loss functions to realize weakly supervised learning by using a minority of acoustic pressure of systems as labeled solutions. Therefore, the MPIPN can identify the computational conditions including different domains and parametric conditions of the acoustic-structure systems. The well-trained MPIPN is capable of solving new unseen parametric conditions in acoustic-structure systems. We applied the MPIPN to parametric steady acoustic-structure systems to validate the effectiveness of the framework and obtained the following results:

(1) The MPIPN is capable of identifying separate constant explicit and implicit parametric conditions for the Helmholtz equations that govern the acoustic-structure systems. Our method predicts solutions of the parametric acoustic-structure systems with constant frequency and physical parameters condition at average relative errors less than $1.0\%$.

(2) The MPIPN is able to deal with changeable unseen combinations of parametric conditions for the Helmholtz equations. The proposed method realizes achieving average relative domain errors of less than $2.82\%$ and $2.21\%$ under unseen combinations of changeable frequencies and physical parameters, respectively.

(3) The MPIPN is efficient and robust for solving parametric acoustic-structure systems with multiple computational domains. By using the proposed method, average relative domain errors of predicted solutions in the acoustic-structure system are less than $2.1\%$ and maximum error is less than $5.0\%$ across all three computational domains.

(4) The MPIPN effectively leverages physics-informed impact to predict pointwise solutions of the acoustic-structure system. The ablation experiment extensively illustrates that the incorporation of the physics-informed term in the loss function results in average improvement over $78\%$ in the accuracy, thereby demonstrating its necessary efficacy.

We plan to broaden the MPIPN in mainly three ways: (1) one is that more types of physics fields and physical knowledge should be involved orientally in the framework more than acoustic-structure systems, (2) another is that a more general method is required to deal with the degenerate solutions for complex PDEs to avoid necessity of any priori solution, (3) the last one is to generalize the feature extraction to the unit cell scale of the model to capture minor perturbations.

# References

[1] G. Liao, Z. Wang, C. Luan, J. Liu, X. Yao, J. Fu, Broadband controllable acoustic focusing and asymmetric focusing by acoustic metamaterials, Smart Materials and Structures, 30 (2021) 045021.

[2] B.-I. Popa, L. Zigoneanu, S.A. Cummer, Experimental acoustic ground cloak in air, Physical review letters, 106 (2011) 253901.

[3] G.d.N. Almeida, E.F. Vergara, L.R. Barbosa, A. Lenzi, R.S. Birch, Sound absorption metasurface with symmetrical coiled spaces and micro slit of variable depth, Applied Acoustics, 183 (2021) 108312.

[4] A.K. Sahai, M. Kaur, S. Joseph, A. Dey, R. Phani, R. Mandal, R. Chattopadhyay, Multi-model multi-physics ensemble: a futuristic way to extended range prediction system, Frontiers in Climate, 3 (2021) 655919.

[5] M. Renardy, R.C. Rogers, An introduction to partial differential equations, Springer Science & Business Media, 2006.

[6] M. Raissi, P. Perdikaris, G.E. Karniadakis, Physics-informed neural networks: A deep learning framework for solving forward and inverse problems involving nonlinear partial differential equations, Journal of Computational physics, 378 (2019) 686-707.

[7] S. Cai, Z. Wang, S. Wang, P. Perdikaris, G.E. Karniadakis, Physics-informed neural networks for heat transfer problems, Journal of Heat Transfer, 143 (2021) 060801.

[8] E. Haghighat, M. Raissi, A. Moure, H. Gomez, R. Juanes, A physics-informed deep learning framework for inversion and surrogate modeling in solid mechanics, Computer Methods in Applied Mechanics and Engineering, 379 (2021) 113741.

[9] Q. Lou, X. Meng, G.E. Karniadakis, Physics-informed neural networks for solving forward and inverse flow problems via the Boltzmann-BGK formulation, Journal of Computational Physics, 447 (2021) 110676.

[10] G.E. Karniadakis, I.G. Kevrekidis, L. Lu, P. Perdikaris, S. Wang, L. Yang, Physics-informed machine learning, Nature Reviews Physics, 3 (2021) 422-440.

[11] X. Meng, Z. Li, D. Zhang, G.E. Karniadakis, PPINN: Parareal physics-informed neural network for time-dependent PDEs, Computer Methods in Applied Mechanics and Engineering, 370 (2020) 113250.

[12] J. Wu, X. Feng, X. Cai, X. Huang, Q. Zhou, A deep learning-based multi-fidelity optimization method for the design of acoustic metasurface, Engineering with Computers, (2022) 1-19.

[13] C.R. Qi, H. Su, K. Mo, L.J. Guibas, Pointnet: Deep learning on point sets for 3d classification and segmentation, in: Proceedings of the IEEE conference on computer vision and pattern recognition, 2017, pp. 652-660.

[14] A. Kashefi, D. Rempe, L.J. Guibas, A point-cloud deep learning framework for prediction of fluid flow fields on irregular geometries, Physics of Fluids, 33 (2021).

[15] P.L. Lagari, L.H. Tsoukalas, S. Safarkhani, I.E. Lagaris, Systematic construction of neural forms for solving partial differential equations inside rectangular domains, subject to initial, boundary and interface conditions, International Journal on Artificial Intelligence Tools, 29 (2020) 2050009.

[16] N. Gao, M. Wang, B. Cheng, Deep auto-encoder network in predictive design of Helmholtz resonator: on-demand prediction of sound absorption peak, Applied Acoustics, 191 (2022) 108680.

[17] B. Wang, Zhang, W. & Cai, W., Multi-Scale Deep Neural Network (MscaleDNN) Methods for Oscillatory Stokes Flows in Complex Domains, Communications in Computational Physics, 28 (2020) 2139-2157.

[18] Z. Liu, W. Cai, Z.-Q.J. Xu, Multi-scale deep neural network (MscaleDNN) for solving Poisson-Boltzmann equation in complex domains, arXiv preprint arXiv:2007.11207, (2020).
[19] J. Sirignano, K. Spiliopoulos, DGM: A deep learning algorithm for solving partial differential equations, Journal of computational physics, 375 (2018) 1339-1364.
[20] G. Kissas, Y. Yang, E. Hwuang, W.R. Witschey, J.A. Detre, P. Perdikaris, Machine learning in cardiovascular flows modeling: Predicting arterial blood pressure from non-invasive 4D flow MRI data using physics-informed neural networks, Computer Methods in Applied Mechanics and Engineering, 358 (2020) 112623.
[21] Y. Zhu, N. Zabaras, P.-S. Koutsourelakis, P. Perdikaris, Physics-constrained deep learning for high-dimensional surrogate modeling and uncertainty quantification without labeled data, Journal of Computational Physics, 394 (2019) 56-81.
[22] N. Geneva, N. Zabaras, Modeling the dynamics of PDE systems with physics-constrained deep auto-regressive networks, Journal of Computational Physics, 403 (2020) 109056.
[23] J.S. Hesthaven, G. Rozza, B. Stamm, Certified reduced basis methods for parametrized partial differential equations, Springer, 2016.
[24] J. Im, F.P. de Barros, S.F. Masri, Data-driven identification of partial differential equations for multi-physics systems using stochastic optimization, Nonlinear Dynamics, 111 (2023) 1987-2007.
[25] J. Im, F.P. de Barros, S. Masri, M. Sahimi, R.M. Ziff, Data-driven discovery of the governing equations for transport in heterogeneous media by symbolic regression and stochastic optimization, Physical Review E, 107 (2023) L013301.
[26] J. Berner, M. Dablander, P. Grohs, Numerically solving parametric families of high-dimensional Kolmogorov partial differential equations via deep learning, Advances in Neural Information Processing Systems, 33 (2020) 16615-16627.
[27] Y. Chen, B. Dong, J. Xu, Meta-mgnet: Meta multigrid networks for solving parameterized partial differential equations, Journal of computational physics, 455 (2022) 110996.
[28] J. Gasick, X. Qian, Isogeometric neural networks: A new deep learning approach for solving parameterized partial differential equations, Computer Methods in Applied Mechanics and Engineering, 405 (2023) 115839.
[29] H. Gao, L. Sun, J.-X. Wang, PhyGeoNet: Physics-informed geometry-adaptive convolutional neural networks for solving parameterized steady-state PDEs on irregular domain, Journal of Computational Physics, 428 (2021) 110079.
[30] A. Kashefi, T. Mukerji, Physics-informed PointNet: A deep learning solver for steady-state incompressible flows and thermal fields on multiple sets of irregular geometries, Journal of Computational Physics, 468 (2022) 111510.
[31] L. Lu, P. Jin, G. Pang, Z. Zhang, G.E. Karniadakis, Learning nonlinear operators via DeepONet based on the universal approximation theorem of operators, Nature machine intelligence, 3 (2021) 218-229.
[32] L. Lu, X. Meng, S. Cai, Z. Mao, S. Goswami, Z. Zhang, G.E. Karniadakis, A comprehensive and fair comparison of two neural operators (with practical extensions) based on fair data, Computer Methods in Applied Mechanics and Engineering, 393 (2022) 114778.
[33] S. Goswami, M. Yin, Y. Yu, G.E. Karniadakis, A physics-informed variational DeepONet for predicting crack path in quasi-brittle materials, Computer Methods in Applied Mechanics and Engineering, 391 (2022) 114587.

[34] Z. Li, N.B. Kovachki, K. Azizzadenesheli, K. Bhattacharya, A. Stuart, A. Anandkumar, Fourier Neural Operator for Parametric Partial Differential Equations, in: International Conference on Learning Representations, 2020.
[35] T. Tripura, S. Chakraborty, Wavelet neural operator for solving parametric partial differential equations in computational mechanics problems, Computer Methods in Applied Mechanics and Engineering, 404 (2023) 115783.
[36] J. Feng, X. Zheng, H. Wang, H. Wang, Y. Zou, Y. Liu, Z. Yao, Low-frequency acoustic-structure analysis using coupled FEM-BEM method, Mathematical Problems in Engineering, 2013 (2013).
[37] D. Misra, Mish: A self regularized non-monotonic activation function, arXiv preprint arXiv:1908.08681, (2019).
[38] F. Zangeneh-Nejad, R. Fleury, Active times for acoustic metamaterials, Reviews in Physics, 4 (2019) 100031.
[39] H. Song, X. Ding, Z. Cui, H. Hu, Research progress and development trends of acoustic metamaterials, Molecules, 26 (2021) 4018.
[40] N. Yu, P. Genevet, M.A. Kats, F. Aieta, J.-P. Tetienne, F. Capasso, Z. Gaburro, Light propagation with phase discontinuities: generalized laws of reflection and refraction, science, 334 (2011) 333-337.
[41] D. Givoli, B. Neta, High-order non-reflecting boundary scheme for time-dependent waves, Journal of Computational Physics, 186 (2003) 24-46.
[42] L. Liu, H. Jiang, P. He, W. Chen, X. Liu, J. Gao, J. Han, On the variance of the adaptive learning rate and beyond, arXiv preprint arXiv:1908.03265, (2019).
[43] M. Zhang, J. Lucas, J. Ba, G.E. Hinton, Lookahead optimizer: k steps forward, 1 step back, Advances in neural information processing systems, 32 (2019).